\documentclass[sigconf]{acmart}
\usepackage[utf8]{inputenc} 
\usepackage[T1]{fontenc}    
\usepackage{hyperref}       
\usepackage{url}            
\usepackage{booktabs}       
\usepackage{amsfonts}       
\usepackage{nicefrac}       
\usepackage{microtype}      
\usepackage{xcolor}         

\usepackage{comment}
\usepackage[american]{babel}

\usepackage{mathtools} 

\usepackage{booktabs} 
\usepackage{tikz} 
\usetikzlibrary{shapes,arrows,positioning,automata}
\usetikzlibrary{decorations.pathreplacing,calligraphy}

\usepackage{mathtools}
\usepackage{cuted}
\usepackage{caption}
\usepackage{subcaption}
\usepackage{multirow}
\usepackage{color,soul}
\usepackage[ruled,linesnumbered]{algorithm2e}
\newtheorem{theorem}{Theorem}
\newtheorem{lemma}{Lemma}

\newcommand{\secfloat}{\textsf{SecFloat}}
\newcommand{\ezpc}{\textsf{EzPC}}
\newcommand{\forward}{\textsf{F-\secfloat}}
\newcommand{\backward}{\textsf{B-\secfloat}}
\newcommand{\backwardl}{\textsf{BL-\secfloat}}

\AtBeginDocument{%
  \providecommand\BibTeX{{%
    \normalfont B\kern-0.5em{\scshape i\kern-0.25em b}\kern-0.8em\TeX}}}

\setcopyright{acmlicensed}
\copyrightyear{2018}
\acmYear{2023}
\acmDOI{XXXXXXX.XXXXXXX}
\acmConference[PP-MAS]{Privacy Preserving Multi Agent Systems}{2023}{USA}

\begin{document}

\title{Privacy Preserving Multi-Agent Reinforcement Learning in Supply Chains}

\author{Ananta Mukherjee}
\email{t-mukherjeea@microsoft.com}
\affiliation{%
  \institution{Microsoft Research India\\
  Microsoft}
  \country{India}
}

\author{Peeyush Kumar}
\email{peeyush.kumar@microsoft.com}
\affiliation{%
  \institution{Microsoft Research Redmond\\
  Microsoft}
  \country{USA}
}

\author{Boling Yang}
\email{bolingy@cs.washington.edu}
\affiliation{%
  \institution{University of Washington Seattle}
  \country{USA}
}

\author{Nishanth Chandran}
\email{nichandr@microsoft.com}
\affiliation{%
  \institution{Microsoft Research Redmond\\
  Microsoft}
  \country{India}
}

\author{Divya Gupta}
\email{divya.gupta@microsoft.com}
\affiliation{%
  \institution{Microsoft Research Redmond\\
  Microsoft}
  \country{India}
}

\begin{abstract}
This paper addresses privacy concerns in multi-agent reinforcement learning (MARL), specifically within the context of supply chains where individual strategic data must remain confidential. Organizations within the supply chain are modeled as agents, each seeking to optimize their own objectives while interacting with others. As each organization's strategy is contingent on neighboring strategies, maintaining privacy of state and action-related information is crucial. To tackle this challenge, we propose a game-theoretic, privacy-preserving mechanism, utilizing a secure multi-party computation (MPC) framework in MARL settings. Our major contribution is the successful implementation of a secure MPC framework, \secfloat\ on \ezpc\, to solve this problem. However, simply implementing policy gradient methods such as MADDPG operations using \secfloat\, while conceptually feasible, would be programmatically intractable. To overcome this hurdle, we devise a novel approach that breaks down the forward and backward pass of the neural network into elementary operations compatible with \secfloat\ , creating efficient and secure versions of the MADDPG algorithm. Furthermore, we present a learning mechanism that carries out \textit{floating point} operations in a privacy-preserving manner, an important feature for successful learning in MARL framework. Experiments reveal that there is on average 68.19\% less supply chain wastage in 2 PC compared to no data share, while also giving on average 42.27\% better average cumulative revenue for each player. This work paves the way for practical, privacy-preserving MARL, promising significant improvements in secure computation within supply chain contexts and broadly.
\end{abstract}









\maketitle 

\section{Introduction}\label{sec:intro}
Supply chains are integral to modern commerce, ensuring the smooth transition of goods and services from production to the consumer. However, operating an efficient supply chain necessitates close coordination and decision-making among multiple organizations. In this context, each organization must often calibrate its strategies based on those of others to ensure effective operational functioning. Despite the clear advantages of such close coordination, it presents significant challenges. One of the primary issues concerns the sharing of private data between organizations. This data is vital for ensuring efficiency, as evidenced by the multi-agent reinforcement learning (MARL) model proposed by \citet{Zhang2021} and further explored by \citet{li2022general} and \citet{Lowe2017MultiAgentAF}. Their research showed improved outcomes in a supply chain game when parties shared states and actions. However, in practice, many supply chain organizations are reluctant to share private data, fearing potential security risks, the possibility of unfair competitive practices, and potential regulatory and legal issues \citep{colicchia2019information}. This reluctance forms a significant barrier to the full realization of the benefits of data-driven decision-making in supply chain operations. Given this context, our research aims to bridge this gap by addressing a critical question: How can we facilitate effective data sharing in supply chains without compromising privacy and security?

To address this problem, the paper propose a solution that utilizes a novel adaptation of secure multi-party computation (MPC) within a multi-agent reinforcement learning (MARL) framework. In our model, we consider a simplified scenario involving a 2-player, where each player wants to optimize towards their objectives separately. We use multi-agent deep deterministic policy gradient (MADDPG) method to simulate this scenario and derive optimal strategies. This approach seeks to enhance supply chain efficiency while preserving data privacy, thereby addressing the prevalent concerns over data sharing in supply chain organizations. Secure multi-party computation (MPC; \cite{evans2018pragmatic}) allows parties to collaboratively perform computations on their combined data sets without revealing the data they possess to each other. We develop secure 32-bit single-precision floating-point gadgets for MADDPG based on SECFLOAT framework by \citet{9833697}, a secure 2PC framework for 32-bit single-precision floating-point operations and math functions. To aid easy development we use \ezpc\ a high-level C-like language that allows developers to express functions to be computed using 2PC (2 Party Computation) without requiring the developer to have any cryptographic knowledge.

It is conceptually possible to execute every primitive arithmetic operation of the underlying MADDPG scheme via equivalent operations of any secure 2PC framework and obtain a privacy preserving MARL framework. However, implementing every primitive arithmetic operation of the underlying MADDPG scheme via equivalent operations of any secure 2PC framework becomes intractable due to the complexity of machine learning programming and high-level maintenance (see \citet{cryptoeprint:2020/300}). Overcoming this hurdle, we devise a novel approach. We break down the forward and backward pass of the neural network into elementary operations. These operations are compatible with the \secfloat\ API, thus creating efficient and secure versions of the MADDPG algorithm. These methods include a forward pass gadget (F-\secfloat\ ) and a backpropagation gadget (B-\secfloat\ ), allowing for secure computation of policy gradient methods such as MADDPG. These gadgets are built for a floating-point model enabling learning using MADDPG.
The generalization of this work beyond supply chains and 2 player setting is discussed in Section~\ref{sec:Generalizations}. We believe this work sets a new course for practical, privacy-preserving MARL and presents substantial improvements in secure computation. As such, this framework enables previously unavailable data sharing solutions, marking a significant contribution to both the field of privacy-preserving machine learning and application research related to industrial operations.

\subsection{Related Work}
Privacy-preserving machine learning is a field that aims to develop machine learning algorithms that can train models on sensitive data without compromising privacy. Common approaches include differential privacy, which adds noise to data to preserve privacy\citep{Geyer2017DifferentiallyPF, Stan2022ASF}, and federated learning \citep{Li2022WindPF}, which trains models on local devices and only shares model updates, not raw data. However, \citet{Bagdasaryan2018HowTB} showed how attackers can potentially add hidden back doors in these models to make it behave in a desired way. Our proposed MPC-based method offers a different approach that allows for direct computation on encrypted data. Secure MPC unlocks a variety of machine-learning applications that are currently infeasible because of data privacy concerns. Although a range of studies implements machine-learning models via secure MPC, mainstream adoption is relatively low. Application to Secure MPC includes domain-specific tasks such as in medical research \citep{Dugan2016ASO} or studying the gender gap in organization salaries \citep{Lapets2018AccessiblePW}. \citet{juvekar2018gazelle, riazi2018chameleon, tan2021cryptgpu} study rigorous privacy and security guarantees for secure MPC frameworks for machine learning tasks. Some work achieved secure deep learning (majorly inference with some scope for training) by converting floating-point models to fixed-point models \citep{Kumar2019CrypTFlowST, Mohassel2017SecureMLAS, Chandran2021SIMCMI, Boemer2020MP2MLAM, Knott2021CrypTenSM, Mishra2020DelphiAC, Liu2017ObliviousNN, Patra2020ABY20IM}. This is not suitable for learning in large or complex MARL setup because of the compounding of errors involved in the conversion from floating point to fixed point. Considering broadly, the adoption of secure MPC is hampered by the complexity of secure MPC and cryptographic techniques for machine learning scientists \& engineers, see \cite{knott2021crypten}.\\

\textbf{MPC in RL:} The computational intricacies inherent in reinforcement learning (RL) methods pose additional challenges when attempting to integrate secure MPC techniques. This research for using a secure MPC framework with RL methods is far and few in between. \citet{Sakuma2008PrivacypreservingRL} showed that the on-line learning method SARSA (state-action-reward-state-action) can be implemented in a privacy-preserving manner. \citet{miyajima2017proposal} proposes a secure MPC framework for Q-learning methods. Yet, both SARSA and Q-learning are RL algorithms that  are designed for single-agent systems, and they are insufficient for applications in a multi-agent setting such as pricing and inventory management in a supply chain system. Unlike these frameworks, the MADDPG framework \citep{lowe2017multi} used here solves MARL where each agent has their own objectives and with no exchange of private information in the clear. There has been more work using federated learning approaches for privacy preservation in reinforcement learning applications \citep{lin2023privacy, li2023wind}. Federated learning allows multiple actors to learn a common model from local datasets without explicitly exchanging training data samples. Unlike federated learning in RL, which require a universal trustworthy server, our method allows the MARL process to stably converge to a high-quality equilibrium through a secure usage of shared information, yet none of the agents have access to the true information of others since all shared data is encrypted. 

\subsection{Key Technical Contributions}\label{subsec:contributions}
This pioneering work addresses the privacy-preserving Markov game problem in a secure multi-party computation (MPC) framework using a policy gradient learning mechanism. Key contributions include:
\begin{itemize}
\item \textbf{New \secfloat\ gadgets:} A novel approach that simplifies the input handling and learning processes of the neural network into basic operations, which can be executed using the \secfloat\ API. These methods include a forward-pass gadget (F-\secfloat\ ) and a backpropagation gadget (B-\secfloat\ ). We show that these gadgets have a significant less computational overhead than native implementation of \secfloat\ . Additionally from a programmatic perspective, we also show that  the development of these gadgets makes it feasible to implement a secure MPC framework for MADDPG setting. Thus, this technique enables secure computation of policy gradient methods like MADDPG, overcoming limitations of previous methods and making this approach privacy-preserving, computationally feasible, and more widely accessible to developers.
\item \textbf{End-to-end Secure Learning Faith-fullness:} The \backward\ module facilitates privacy-preserving gradient optimization in the presence of strategic agents using MADDPG. Our floating-point secure model based on \secfloat\  preserves accuracy of the learning mechanism suitable for large number iterations of the backward pass. Unlike previous fixed-point models, we empirically show that the mechanism involving both inferences and gradient updates during learning remains end-to-end faithful in the secure universe compared to the clear-text analogue. 
\item \textbf{Efficacy Proven by Experiments:} Testing in a 2-player supply chain setting demonstrated resource wastage reduced by an average of 68.19\% and improved average cumulative revenue by 42.27\% for each player when compared to no data sharing systems. By significantly reducing wastage and improving revenue in industries involving multi-party decision making, these findings could lead to more efficient and secure operations, thereby benefiting the economy and society as a whole.
\end{itemize}

\section{Background: Cryptographic Primitives}\label{CryptographicPrimitives}
Secure Multi-party Computation (MPC) is a protocol that facilitates joint computation over private inputs from multiple parties. Think of it as a method that allows companies to collaborate on optimizing revenue without revealing individual data. Our research uses MPC to enable agents to learn from others while keeping their own information private. Suppose we have parties $P_1,P_2,...,P_n$ each with secret inputs $x_1,x_2,...,x_n$. They want to compute a function $f(x_1,x_2,...,x_n)$ together. The ideal scenario would involve a universally trusted party $T$ that calculates $f$ and shares the output with all parties, but nothing more. However, such a trusted entity often doesn't exist in real-life scenarios. Cryptography resolves this with MPC, which facilitates a protocol that involves regular information exchange. This protocol ensures \textbf{Correctness} (output as if computed by $T$) and \textbf{Input Privacy} (no additional information beyond the output is disclosed). For more information, see \citet{cryptoeprint:2020/300}.
\\\noindent {\bf Secret sharing:} A $(t+1)$-out-of-$n$ secret sharing scheme is a method by which a dealer can distribute a secret $s$ amongst $n$ parties. This scheme is designed in such a way that any subset of $(t+1)$ or more parties can reconstruct the secret uniquely by combining their shares, but any subset of $t$ or fewer parties can't glean anything about the secret \citep{Shamir1979HowTS}. Please refer to the supplementary material and \citet{10.1145/62212.62213} for more details.
\\\noindent{\bf Invariants in MPC:} An important feature of most MPC protocols is their ability to maintain an invariant: the parties can start with secret shares of values $x_1, ..., x_n$ and execute a protocol such that they end up with secret shares of $f(x_1, x_2, \cdots, x_n)$. This invariant is maintained for several simple functions. Once this is available for a set of functions $f_1, \cdots, f_k$, these functions can be composed to compute the composition of these functions.
\\\textbf{\secfloat\ and \ezpc:} When it comes to secure training of machine learning models, we often need to compute functions such as addition, multiplications, comparisons, division, tangent, and so on. While theoretically, all functions can be expressed in terms of additions and multiplications, doing so securely would be computationally expensive. \secfloat\ \citep{9833697} is a state-of-art framework that performs secure 2PC of 32-bit single-precision floating-point operations and mathematical functions like comparison, addition, multiplication, division, and other transcendental functions in an efficient manner. \ezpc\ is a user-friendly, high-level C-like language that allows developers to express functions to be computed using 2PC, particularly using \secfloat, without requiring the developer to have deep knowledge of cryptography. The \ezpc\ compiler generates a secure computation protocol for the entire function based on \secfloat. The \secfloat\ protocol ensures that the ULP (units in last place) errors between floating-point values obtained by elementary \secfloat\ operations and the corresponding exact real results are less than 1. This result is important to preserve the integrity of our learning algorithm.
\begin{lemma}\label{theorem}[Precision Guarantee \citep{Rathee2022SecFloatAF}]
If $r^{-}$ is the floating-point number just below $r$ and $r^\prime$ is the floating-point number just above $r$, then $ULP(r,r^\prime)=\frac{\vert r-r^\prime\vert}{r^{+}-r^{-}}$,
where $r^\prime$ is the floating point number and $r$ is the real number
\end{lemma}

\begin{figure*}[htbp]
    \centering
\includegraphics[width=0.9\textwidth]{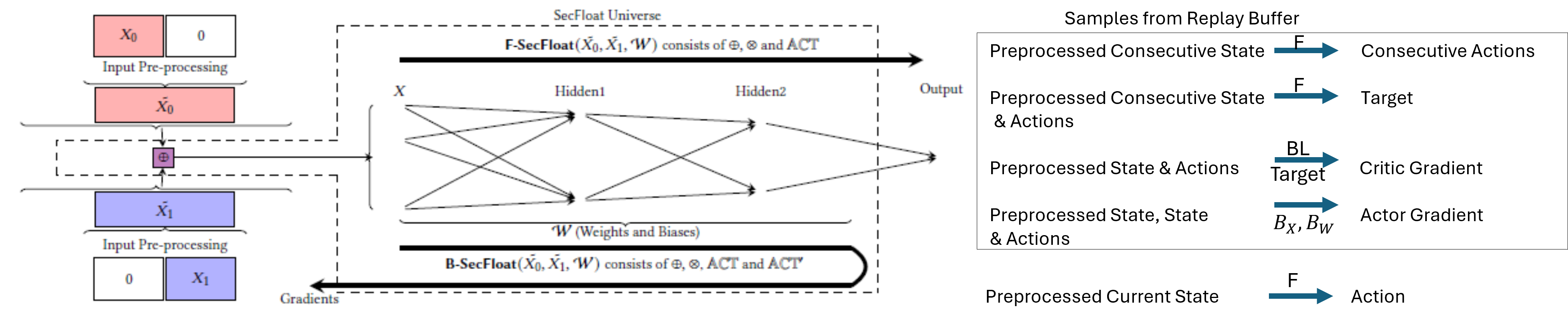}
\caption{{\footnotesize Input Pre-processing and {\secfloat} based Secure Forward and Backward Pass Framework for Player 0. Red and Blue denote private data of Player 0 and encrypted data of Player 1 respectively.$\oplus,\otimes$,$\mathbb{ACT}$ and $\mathbb{ACT}^\prime$ respectively denote {\secfloat} modules for Matrix Addition, Matrix Multiplication, Various Activation Functions (viz ReLU, Sigmoid etc) and derivative of Activation Functions.}}\label{fig:SecureNNModules}
\end{figure*}
\section{\secfloat\  Based Secure Neural Network Computations}\label{SecureForwardAndBackwardPasses}
We will illustrate the development of secure modules on MADDPG using the SECFLOAT API using a three-layer neural network as an example.
Let $N(X\vert\mathcal{W}):\mathbb{R}^{B\times 2d}\rightarrow\mathbb{R}^{B\times z}$ be a 3 layered feed forward neural network with input dimension $2d$, output dimension $z$, hidden dimension $h$ and batch size $B$. The layers are fully connected. $\mathcal{W}=\{W_1\in \mathbb{R}^{h\times 2d},b_1\in \mathbb{R}^{1\times h},W_2\in \mathbb{R}^{h\times h},b_2\in \mathbb{R}^{1\times h},W_3\in \mathbb{R}^{z\times h},b_3\in \mathbb{R}^{1\times z}\}$ where $W_i$ and $b_i$ are the weight matrix and the bias vector, respectively for the $i$'th layer. Let $f_i$ denote the activation function for the $i$'th layer, and let $b_i^B$ represent a matrix with $B$ rows where each row is $b_i$. The forward pass for N is
\begin{equation}\label{ReLuNN}
\resizebox{\columnwidth}{!}{%
$N(X)=\underbrace{f_3}_{layer3Act}\underbrace{\overbracket{\Biggl[\biggl(\underbrace{f_2}_{layer2Act}\underbrace{\Bigl[\overbracket{\bigl(\underbrace{f_1}_{layer1Act}\underbrace{[\overbracket{X}^{layer1In} W_1^T+b_1^B]}_{layer1Out}\bigr)}^{layer2In}W_2^T+b_2^B\Bigr]}_{layer2Out}\biggr)}^{layer3In}W_3^T+b_3^B\Biggr]}_{layer3Out}$
}
\end{equation}
For layer number $i\in\{1,2,3\}$ we define: $layer-i-In:$ Input of layer "i", $layer-i-Out:$ Output of layer "i", $layer-i-Act:$ Activation Function on top of layer "i", $layer-i-Der:$ Gradient of $N(X)$ with respect to the output of layer "i" and $layer-i-ActDer:$ Gradient of $N(X)$ with respect to the output of the activation function on top of layer "i". For $i\in\{1,2,3\}$, $layer-i-In$, $layer-i-Out$ and $layer-i-Act$ are shown in equation~\ref{ReLuNN}. 

Within a 2-party setting, our objective is to construct two modules using the \secfloat\ framework: the forward-pass gadget (F-\secfloat\ API) and the backpropagation gadget (B-\secfloat\ API). This requires preprocessing the data and defining certain configurations:
\\\textbf{Input Distribution between 2 Parties: }Let, $x^j$ is the $j$'th row vector of X for $j=1,2,...,B$. Each input vector in the batch, represented as $x^j$, is divided into two halves. These halves serve as secret vectors for Party 0 and Party 1 respectively, meaning $x^j=x^j_0:||:x^j_1$ where $x^j_i\in \mathbb{R}^{d}$ for $i=1,2$. Here $||$ is the vector concatenation operator. $
    X_0 = [x^i_0 \:\forall i \in [1,B]]_{B\times d};
     X_1 = [x^i_1 \: \forall i \in [1,B]]_{B\times d}
$ represent the secret matrices held by party 0 and 1.  Effectively $X=X_0\:\bigr]\bigl[\:X_1$, where $\mathbf{\bigr]\bigl[}$ is the row wise matrix concatenation operator. A specific Party $i$ holds all the weights and biases of the neural network i.e., $\mathcal{W}=\mathcal{W}_i$. 
\\\textbf{Output: }Party $i$ should receive both the forward pass prediction, represented as $N(X\vert\mathcal{W})$, and the desired gradient for the backward pass. In contrast, Party $(1-i)$ should not receive anything. 
\\\textbf{Adversarial model: }Semi-honest i.e., the parties don't deviate from the prescribed protocols but they are keen to tap sensitive information of the other. \secfloat\ provides security against a static probabilistic polynomial time semi-honest adversary.

\subsection{High-level APIs for Privacy-preserving Forward and Backward Passes}\label{sec:API's}
\textbf{Input Pre-processing: }For every input vector in the batch, the first half is possessed by the Party 0 and the second half is possessed by Party 1, they need to pre-process the input matrices to make them compatible for the secure forward and backward pass APIs. Here on, we denote the pre-processed matrices by an overhead tilde sign. Party 0 and 1 prepare respectively, $ \Tilde{X_0}=X_0\:\bigr]\bigl[\:0^{B\times d}, \Tilde{X_1}=0^{B\times d}\:\bigr]\bigl[\:X_1 $. Here, $0^{B\times d}$ is an all-zero matrix. We precisely get, $(\Tilde{X_0}+\Tilde{X_1})=X$. Left Part of figure~\ref{fig:SecureNNModules} captures this.
\\\textbf{Construction Details of the APIs:}
Next, we will delineate the construction of secure forward and backward pass APIs, designed to be executed by each player. With \secfloat\ ensuring the security of basic mathematical operations, functions, and their compositions, we can combine them in a hierarchical manner to develop secure high-level functionalities. 
\\\textbf{For Forward Pass: }For our current setting, without loss of generality, let in equation ~\ref{ReLuNN}, $f_1$ and $f_2$ are ReLU and $f_3$ is identity, then:
\begin{theorem}\label{ftheorem}
Let $\otimes$ and $\oplus$ denote the \secfloat\ APIs for matrix multiplication and addition respectively. Denote the ReLU API in \secfloat\ by $\mathbb{RELU}$. The forward-pass gadget, with respect to equation~\ref{ReLuNN},
\begin{equation}\label{F-SecFloat}
\resizebox{\columnwidth}{!}{%
$\forward(\Tilde{X_0},\Tilde{X_1},\mathcal{W})=\Bigl[\mathbb{RELU}\Bigl(\bigl[\mathbb{RELU}\bigl((\Tilde{X_0}\oplus\Tilde{X_1})\otimes W_1^T\oplus b_1\bigr)\bigr]\otimes W_2^T\oplus b_2\Bigr)\Bigr]\otimes W_3^T\oplus b_3$
}
\end{equation}
\end{theorem}
Thus  $\mathbf{\textbf{\forward}(\Tilde{X_0},\Tilde{X_1},\mathcal{W})}$ provides an abstraction for forward-pass in a secure manner.
\\\textbf{For Backward Pass:} Before jumping into the gradient expressions we define 2 things. The gradient of $ReLU$ as $ReLU^\prime(x)$.
A function $GetBiasDer$ that will help us to calculate gradients with respect to biases:
\begin{equation}\label{GetBiasDer}
    GetBiasDer(M^{B\times n})=\Bigl[\sum_{i=1}^BM_{i1},\:...,\:\sum_{i=1}^BM_{ij},\:...,\:\sum_{i=1}^BM_{in}\Bigr]^{1\times n}
\end{equation}
\secfloat\ supports for $ReLU^\prime$ and $GetBiasDer$ are available. Let $\circ$ denotes element wise product. By chain rule, we have the following system:

\begin{footnotesize}
\begin{align}
& layer3Der=\frac{1^{B\times z}}{B.z} \label{layer3Der}\\
& \nabla_{W_3}N=(layer3Der)^T.(layer3In) \label{GradW3}\\
& \nabla_{b_3}N=GetBiasDer(layer3Der) \label{GradB3}\\
& layer2ActDer=(layer3Der).W_3 \label{layer2ActDer}\\
& layer2Der=ReLU^\prime(layer2Out)\circ(layer2ActDer) \label{layer2Der}\\
& \nabla_{W_2}N=(layer2Der)^T.(layer2In) \label{GradW2} \\
& \nabla_{b_2}N=GetBiasDer(layer2Der) \label{GradB2}\\
& layer1ActDer=(layer2Der).W_2 \label{layer1ActDer}\\
& layer1Der=ReLU^\prime(layer1Out)\circ(layer1ActDer) \label{layer1Der}\\
& \nabla_{W_1}N=(layer1Der)^T.(layer1In) \label{GradW1}\\
& \nabla_{b_1}N=GetBiasDer(layer1Der) \label{GradB1}\\
& \nabla_{X}N=(layer1Der).W_1 \label{GradX}
\end{align}
\end{footnotesize}

\begin{theorem}\label{btheorem}
Following equations~\ref{GradW1}, ~\ref{GradB1}, ~\ref{GradW2}, ~\ref{GradB2}, ~\ref{GradW3}, ~\ref{GradB3} and evaluating all intermediate entities by secure \secfloat\ APIs, we get the first kind of back propagation gadget $\mathbf{\textbf{\backward}_{\mathcal{W}}(\Tilde{X_0},\Tilde{X_1},\mathcal{W})}$ that is the privacy preserving analogue of $\nabla_{\mathcal{W}}N$. Equation~\ref{GradX} gives the gadget  $\mathbf{\textbf{\backward}_{\Tilde{X_i}}(\Tilde{X_0},\Tilde{X_1},\mathcal{W})}$ analogous to $\nabla_{\Tilde{X_i}}N$. For $\nabla_{\mathcal{W}}L(N,t)$ where $t$ is the target and $L$ is the loss function, we get \\$\mathbf{\textbf{\backwardl}_{\mathcal{W}}(\Tilde{X_0},\Tilde{X_1},\mathcal{W},t)}$ by substituting the right hand side of equation~\ref{layer3Der} by $\nabla_{layer3Out}\:L(layer3Out, t)$ followed by using the equations ~\ref{GradW1}, ~\ref{GradB1}, ~\ref{GradW2}, ~\ref{GradB2}, ~\ref{GradW3}, ~\ref{GradB3}.
\end{theorem}
Figure~\ref{fig:SecureNNModules} shows the architecture of input distribution, pre-processing, secure computation of the forward and backward passes and outputs. 
\\\textbf{Generalizing to larger networks:} Here we use a 3 layered fully connected neural networks with activation functions ReLU and Sigmoid. While for our case this was sufficient to achieve stable equilibria for the parties, the gadgets allow generalization of the secure 2PC to more complex architectures in terms of size or convolutions or activations. Corresponding forward and backward passes can be achieved using appropriate combination of developed \secfloat\ gadgets in a similar fashion to what we did. Forward pass is straightforward and backward pass modules can be built by carefully analysing the chain rule, using the template defined in equations~\ref{GradW1}, ~\ref{GradB1}, ~\ref{GradW2}, ~\ref{GradB2}, ~\ref{GradW3}, ~\ref{GradB3} and ~\ref{GradX} for back propagation. We omit detailed descriptions for the sake of brevity.  

\section{Secure MADDPG on Supply Chain Game}\label{sec:game}
\citet{li2022general} gives a general sum stochastic game description for an $N$ party networked supply chain. We build on a 2 player supply chain setup to analyze the secured MPC framework. Player 0 purchases from the raw material market and sells to Player 1, while Player 1 sells to the consumer market. \\\textbf{State: }Player $i$'s internal state is $s_i=[c_i,\mu_i,x_i,y_i]$. $c_i\in\mathbb{R}_{+}$ is the purchasing cost per unit product from $i$'s supplier. $\mu_i\in\mathbb{R}_{+}$ is the demand anticipated by Player $i$ from her retailer. The current stock level of Player $i$ is $x_i\in\mathbb{R}_{+}$ and the incoming stock replenishment in the next $l_i$ time steps is $y_i\in\mathbb{R}_{+}^{l_i}$. So, at time $t$, $[y_i(t)]_n$ is the replenishment that arrives at time $t+n$. \\\textbf{Action: }Player $i'$s action is $a_i=[q_i,p_i]$. $q_i\in\mathbb{R}_{+}$ is the quantity of product that Player $i$ decides to purchase from her supplier and $p_i\in\mathbb{R}_{+}$ is the price per unit product that Player $i$ decides to charge from her retailer.
\\\textbf{Realized and Anticipated Demand: }If Player $i$ receives a total order of $D_i$ to supply and $x_i$ is her current stock level, then in the two-player setup, realized demand by Player $i$ is given by $d_i=min(D_i,x_i)$. Anticipated demand is computed on the basis of historical demand trends i.e., $\mu_i(t)=f_i(D_i(0),D_i(1),...,D_i(t-1))$ where $f_i$ is a suitable forecasting method like the ARMAX by \citep{THOMASSEY2010470}
\\\textbf{Stock Level and Replenishment Transitions: }The stock level at time $t+1$ is given by, $x_i(t+1)=x_i(t)-d_i(t)+[y_i(t)]_1$. The incoming stock replenishment transitions are given by $[y_i(t+1)]_k=[y_i(t)]_{k+1}$ for $k=1,2,...,l_i-1$. The last incoming replenishment $[y_i(t+1)]_{l_i}$ is the available supply from Player $i'$s supplier.
\\\textbf{Rewards: } After a specific transition the total reward of Player $i$ is:
\begin{equation}\label{eqn:RewardEqn}
    \begin{multlined}
    r_i(t) = \underbrace{p_i(t)d_i(t)}_{\text{Total Revenue}} \:-\underbrace{c_i(t)z_i(t)}_{\text{Total Purchasing Cost}} 
    \\ - \underbrace{h_i\left(x_i(t)-d_i(t)\right)}_{\text{Holding Cost}} -  \underbrace{w_i\left(D_i(t)-d_i(t)\right)}_{\text{Loss of Goodwill Cost}},
    \end{multlined}
\end{equation}
where $h_i, w_i\in \mathbb{R}_{+}$ are proportionality constants, and $z_i$ is the quantity delivered by Player $i$'s supplier. Player 0 and Player 1 solve for different reward structure. For detailed expression, please see the supplementary material.
\\\textbf{Wastage: } Since Player 0 fully gets her ordered quantity ($q_0$) delivered from the raw materials market, the quantity procured by this supply chain equals $q_0$.  Player 1 delivers a quantity of $d_{c1}$ to the consumer market. So the wastage due to this supply chain can be characterized as $(q_0 - d_{c1})$. For a healthy supply chain, this wastage should be as low as possible.
\\Each player engages in a stochastic asymmetric information game ~\citep{nayyar2013common} through a Markov Decision Processes (MDP). Each player intends to find an optimal policy for an MDP $\left(\mathcal{S}_i,\mathcal{A}_i,P_i,R_i,\gamma_i\right)$. For Player $i$, $\mathcal{S}_i$ is the set of valid states,  $\mathcal{A}_i$ is the set of all valid actions, $P_i$ and $R_i$ are the transition probability and reward functions. $\gamma_i$ is discount factor in time.

\subsection{Secure MADDPG}
\begin{figure*}
\centering
\begin{minipage}{.55\textwidth}
\centering
\includegraphics[height=0.21\textheight, width=\textwidth]{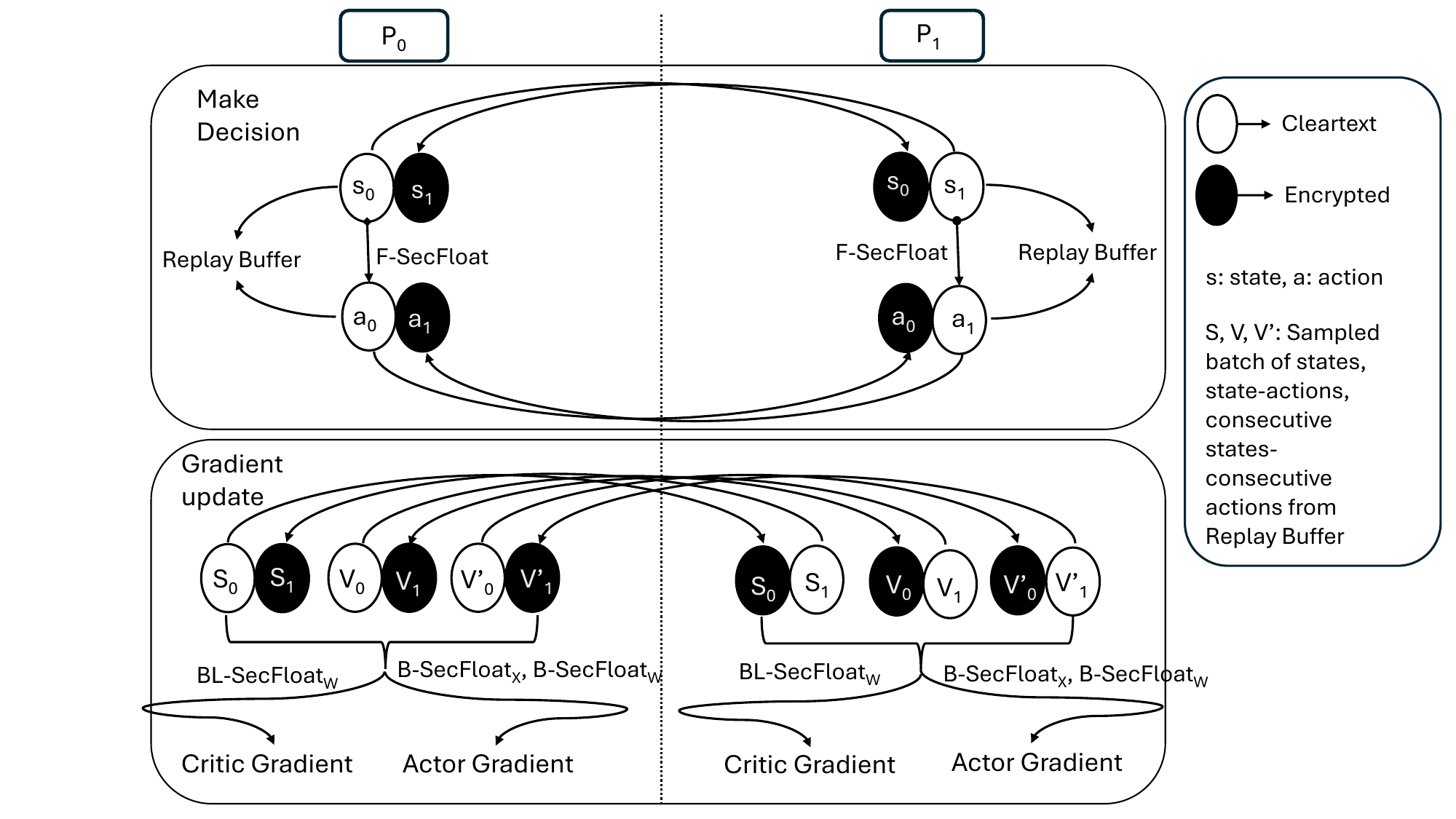}
\captionof{figure}{Mechanism of information sharing when invoking the developed \secfloat\ gadgets.}\label{fig:algorithm-diag}
\Description{Algorithm-diag}
\end{minipage}\qquad
\begin{minipage}{.4\textwidth}
\centering
\includegraphics[height=0.2\textheight, width=\textwidth]{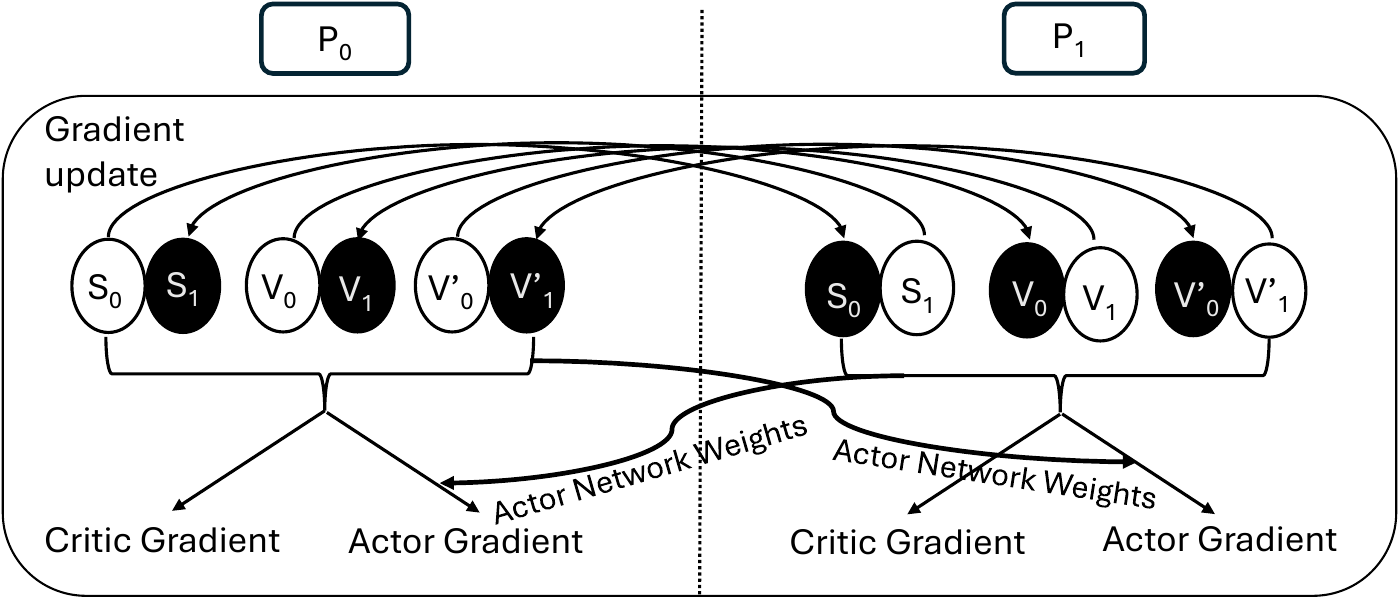}
\captionof{figure}{Gradient update phase of the native algorithm additionally requires sharing of encrypted weights of the other party's Actor Network}\label{fig:naive-scheme}
\Description{naive-schem-diag}
\end{minipage}
\end{figure*}

MADDPG is an extension of Deep Deterministic Policy Gradient (DDPG) \citep{Lillicrap2015ContinuousCW}, which is a policy-based reinforcement learning algorithm. MADDPG differs from DDPG by catering to multi-agent environments where multiple learning agents interact with each other. In a MADDPG setup with mutual information sharing, four neural networks exist for every each Player $i$, namely Actor ($\pi_i$), Critic($Q_i$), Actor Target ($\pi^\prime_i$) and Critic Target ($Q^\prime_i$). They are parameterized by $\theta_i,\phi_i,\theta^\prime_i$ and $\phi^\prime_i$ respectively. Let's consider here only the 2 party version. For Player $i$, apart from her own states and actions $s_i$ and $a_i$, the states and actions of the other player i.e., $s_{1-i}$ and $a_{1-i}$ are also accessible. Under that circumstances, Player $i$ picks up her action according to $a_i=\pi_i(s_i, s_{1-i}\vert\theta_i)$. The Critic Network, $Q_i$ gives the expected reward for Player $i$ if Player $i$ starts from state and action $s_i$ and $a_i$ for $i\in\{0,1\}$ both and then Player $i$ forever follows a trajectory $\tau$ according to her current policy. The expected reward is given by $ Q_i(s_i,a_i,s_{1-i},a_{1-i}\vert\phi_i)
$. Actor Target and Critic Target networks are $\pi^\prime_i(.,.\vert\theta^\prime_i)$ and $Q^\prime_i(.,.,.,.\vert\phi^\prime_i)$ respectively.
When Player $i$ for $i\in\{0,1\}$ both, makes a transition from a state $s_i$ to $s^\prime_i$ by an action $a_i$ with a reward $r_i$, the target expected return is updated independently as
$Q^{targ}_i(r_i,s^\prime_i)=r_i+\gamma Q^\prime_i\Bigl(s^\prime_i,\pi^\prime_i(s^\prime_i,s^\prime_{1-i}\vert\theta^\prime_i),s^\prime_{1-i},\pi^\prime_{1-i}(s^\prime_{1-i},s^\prime_i\vert\theta^\prime_{1-i})\Big|\phi^\prime_i\Bigr).$
\\For a mini batch of transitions, $B_i=\{(s_i,a_i,r_i,s^\prime_i)_k\}$ (for $k=1,2,...,\vert B\vert$) both with the sampling randomness being the same for both $B_0$ and $B_1$, the gradient of the MSBE (Mean squared Bellman Error) is:
\begin{equation}\label{eqn:GradL}
    \begin{multlined}
        \nabla_{\phi_i}L= \frac{2}{|B_i|}\sum_{(s_i,a_i,r_i,s^\prime_i)\sim B_i}\bigl[\bigl(Q_i(s_i,a_i,s_{1-i},a_{1-i}\vert\phi_i)- \\ Q^{targ}_i(r_i,s^\prime_i)\bigr)
        \nabla_{\phi_i}Q_i(s_i,a_i,s_{1-i},a_{1-i}\vert\phi_i)\bigr]
    \end{multlined}
\end{equation}
The policy gradient according to  \citep{Silver2014DeterministicPG} is:
\begin{equation}\label{eqn:PolicyGrad}
    \begin{multlined}
        \nabla_{\theta_i}\frac{1}{|B_i|}\sum_{s_i\sim B_i}Q_i\bigl(s_i,\pi_i(s_i,s_{1-i}\vert\theta_i),s_{1-i},\pi_{1-i}(s_{1-i},s_i\vert\theta_{1-i})\big|\phi_i\bigr)\\
        = \frac{1}{|B_i|}\sum_{s_i\sim B_i}\bigl[\nabla_{a_i}Q_i(s_i,a_i,s_{1-i},a_{1-i}\vert \phi_i)\big|_{a_i=\pi_i(s_i,s_{1-i}\vert\theta_i)}\\
        \nabla_{\theta_i}\pi_i(s_i,s_{1-i}\vert\theta_i)\bigr]
    \end{multlined}
\end{equation}
\\When direct exchange of data is not allowed, then we cannot compute $a_i$, $Q_i$, $Q^{targ}_i$ and Equations ~\ref{eqn:GradL},~\ref{eqn:PolicyGrad} directly. This is where we integrate the privacy preserving \secfloat\ based modules developed in Theorems~\ref{ftheorem} \& \ref{btheorem} of section~\ref{sec:API's} generate hybrid integrated APIs for MADDPG updates. 
\begin{theorem}\label{tmaddpg}
For Party $i$, the updated version of $Q^{targ}_i$ and equations~\ref{eqn:GradL},~\ref{eqn:PolicyGrad} are:
\begin{align}
Q^{targ}_i=&R_i+\gamma\:\forward(\Tilde{V^\prime_i},\widetilde{V^\prime_{1-i}},\phi^\prime_i)\\
    \nabla_{\phi_i}L=&\backwardl_{\phi_i}(\Tilde{V_i},\widetilde{V_{1-i}},\phi_i,Q^{targ}_i) \\
    Policy\: Gradient =&\backward_{A_i}(\Tilde{V_i},\widetilde{V_{1-i}},\phi_i) \circ \backward_{\theta_i}(\Tilde{S_i},\widetilde{S_{1-i}},\theta_i)
\end{align}
\end{theorem}

\SetKwInput{kwInit}{Include}
\begin{algorithm}[htbp]
\caption{An iteration of 2PC with MADDPG \secfloat\ gadgets}\label{alg:Secure}
\begin{small}
\KwIn{ States $s_i$, Actor-network weights $\theta_i$, Critic-network weights $\phi_i$, Target actor-network weights $\theta^\prime_i$, Target critic-network weights $\phi^\prime_i$ - for each Player $i\in\{0,1\}$}
\kwInit{\secfloat\ API's $\forward$, $\backward$, $\backwardl$ as developed in section 3.1}
\For {$i\in\{0,1\}$}{
$\tilde{s_i}\gets$ Input pre-processing of states $s_i$ with Batch Size 1, according to section 3.1\;
Action $a_i\gets clip\bigl(\forward(\tilde{s_i},\widetilde{s_{1-i}},\theta_i)+\epsilon, low, high\bigr)$ where $\epsilon\sim\mathcal{N}(0,1)$ and $low, high$ are bounds\;
$s^\prime_i\gets$ the next state reached due to action $a_i$\;
$r_i\gets$ reward associated to the state transition\;
$(s_i,a_i,r_i,s^\prime_i)$ is stored in $\mathcal{D}_i$\ (replay buffer);
}
Each Player \For {$i\in\{0,1\}$} {
Player $i$ samples an index set $\mathcal{I}_i\subset\vert\mathcal{D}_i\vert$ and sends $\mathcal{I}_i$ explicitly to Player $(1-i)$ where $\vert\mathcal{I}_i\vert$ is the batch size\;
\For {$j\in\{0,1\}$}{
Batch $B_j\gets \{\bigl(s_j,a_j,r_j,s^\prime_j\bigr)_k\in \mathcal{D}_j\}$ where $k\in\mathcal{I}_i$\;
$B\gets \vert B_j\vert=\vert\mathcal{I}_i\vert$\;
$S_j\gets$ stack $(s_j)_1,...,(s_j)_B$'s row wise\;
Similarly stack $A_j$, $R_j$, $S^\prime_j$ row wise\;
$V_j=S_j\bigr]\bigl[ A_j$ (concatenated state-action)\;
$\Tilde{S_j},\Tilde{V_j},\Tilde{S^\prime_j}\gets$ Input pre-processing of $S_j, V_j, S^\prime_j$ according to section 3.1
}
\For {$j\in\{0,1\}$}{
Next Actions: $A^\prime_j\gets\forward(\Tilde{S^\prime_j},\widetilde{S^\prime_{1-j}},\theta^\prime_j)$\;
$V^\prime_j\gets S^\prime_j\bigr]\bigl[ A^\prime_j$ (concatenated next-state, next-action)\; 
$\Tilde{V^\prime_j}\gets$ Input pre-processing of $V_j^\prime$ according to section 3.1\;
}
\textbf{Critic:} \\
\Indp $Q^{targ}_i\gets R_i+\gamma\:\forward(\Tilde{V^\prime_i},\widetilde{V^\prime_{1-i}},\phi^\prime_i)$\ [equation 19]\\
$\nabla_{\phi_i}L\gets\backwardl_{\phi_i}(\Tilde{V_i},\widetilde{V_{1-i}},\phi_i,Q^{targ}_i)$ [equation 20]\;
\Indm
\textbf{Actor:}\\
\Indp $\nabla_a Q\gets\backward_{A_i}(\Tilde{V_i},\widetilde{V_{1-i}},\phi_i)$\\
Policy gradient $\gets\nabla_a Q \circ \backward_{\theta_i}(\Tilde{S_i},\widetilde{S_{1-i}},\theta_i)$\ [$\circ$ is element-wise product. Above 2 steps follow from equation 21]\;
\Indm
$\theta_i\gets\theta_i + lr_a$[Policy gradient]\;
$\phi_i\gets\phi_i - lr_c[\nabla_{\phi_i}L]$, where $lr_a$ and $lr_c$ are actor and critic learning rates\;
}
$\theta_i^\prime \gets \rho\theta_i^\prime + (1-\rho)\theta_i$\;   
$\phi_i^\prime \gets \rho\phi_i^\prime + (1-\rho)\phi_i$. Target network updates at fixed intervals with hyperparameter $\rho$\;
\end{small}
\end{algorithm}
Please see Algorithm~\ref{alg:Secure} and supplementary material for development details of the above theorem and the pseudo-code for an iteration of the secure MADDPG\footnote{Code is made open-source at https://github.com/anonymous-secure-marl/PrivacyPreservingSupplyGame}. 
\subsection{\secfloat\ High-Level Gadgets vs Native Implementation}
It is theoretically possible to execute each arithmetic operation of the MADDPG scheme natively via a secure 2PC, without using the high-level gadgets developed. Yet, implementation becomes infeasible due to machine learning complexity and high-level maintenance (see \citet{cryptoeprint:2020/300} for discussion on limitations of maintenance). The fundamental challenge for the native implementation arises during the backward pass. The agent requires cryptographic information or the secret share from the forward pass of the other agent for every primitive computation. This cross dependence is inherent in equations~\ref{eqn:GradL} and~\ref{eqn:PolicyGrad}. This exchange cannot be reliably achieved as we do not have control over the other agent's computational paradigm and architecture. A single failure in this exchange disrupts the entire MARL computation. Contrarily, when utilizing the developed high-level gadgets, agents' decision-making aligns exactly at crucial iteration stages, avoiding this issue. Using Theorem~\ref{tmaddpg}, one can replace the clear-text forward and backward pass directly with the \forward\ and \backward\ gadgets, enabling the computations to become accessible. For every forward and backward pass we need to pre-process data once and call these gadgets. There are only a constant number of preprocessing in each pass (as shown in line no 2, 16 and 20 in Algorithm~\ref{alg:Secure}). Figure~\ref{fig:naive-scheme} shows the interaction pattern that is needed to be managed during native implementation of the \secfloat\ on MADDPG. We show this for the purpose of illustration but note that it is infeasible to implement this in practice due to reasons highlighted earlier in this section. Contrarily, Figure~\ref{fig:algorithm-diag} shows the interaction pattern with the developed \secfloat\ gadgets which makes the implementation of secure 2PC feasible on MADDPG. 

\textbf{On Timing and Complexity:} Even in the hypothetical scenario where native computations are feasible, the temporal complexity increases quite significantly, lending the native implementation infeasible in practice. Programmatically, the source of this overhead is the pre-processing step. Pre-processing of input data is required everytime before invoking the \secfloat\ APIs for each party. During pre-processing no other computation can happen because of cross-dependency. Theorems~\ref{Preprocessing-theorem-maddpg} and \ref{Preprocessing-theorem-native} formally captures the computational complexity for both cases. The proof is provided in the supplementary material Section F. Given that for MADDPG $f$ is the number of computations for one pass of clear-text without any cryptography computation, $\vert S\vert$ is the state-space size, $\vert A\vert$ the action space size, '$l$' is the number of actor-network nodes, and 'n' the total number of iterations (Number of epochs $\times$ Iterations per epoch):
\begin{theorem}\label{Preprocessing-theorem-maddpg}
    Computational complexity for the entire lifetime of the 2 party secure MADDPG using F-\secfloat\ and B-\secfloat\ gadgets is $\mathcal{O}(f (\vert S\vert + \vert A \vert) n)$.  
\end{theorem}

\begin{theorem}\label{Preprocessing-theorem-native}
Computational complexity for the entire lifetime of the 2 party MADDPG using native \secfloat\ implementation is  $\mathcal{O}(fln)$ in the best case scenario and $\mathcal{O}(f^2 (\vert S\vert + \vert A\vert )n)$ in the worst case scenario. 
\end{theorem}
We see that the naive native implementation exhibits computational complexity that can increase exponentially under worst-case conditions. Even in the best-case scenario, the complexity remains high. Given that we cannot dictate or predict the computational approach and architecture used by the other agent, it's uncertain which scenario will manifest. This unpredictability renders computations in the native scenario impractical.
Additionally, \secfloat\ support both correct floating-point primitive operations (addition, multiplication, division, and comparison) and precise math functions (trigonometric functions, exponentiation, and logarithm), see \citet{Rathee2022SecFloatAF}. This ensures floating point capability for high level secure MADDPG operations, unlike previous privacy preserving mechanisms for MARL settings.
\section{Experiments and Results}
\begin{figure*}[t!]
   \centering
   \caption{"Go-Live" reward, quantity and wastage plots in different modes of execution (Fixed Randomness 1)}
    \label{fig:Randomness1} 
\begin{tabular}{ccc}
Secure 2PC&Explicit Data Exchange&No Data Sharing\\
\includegraphics[width=0.32\textwidth]{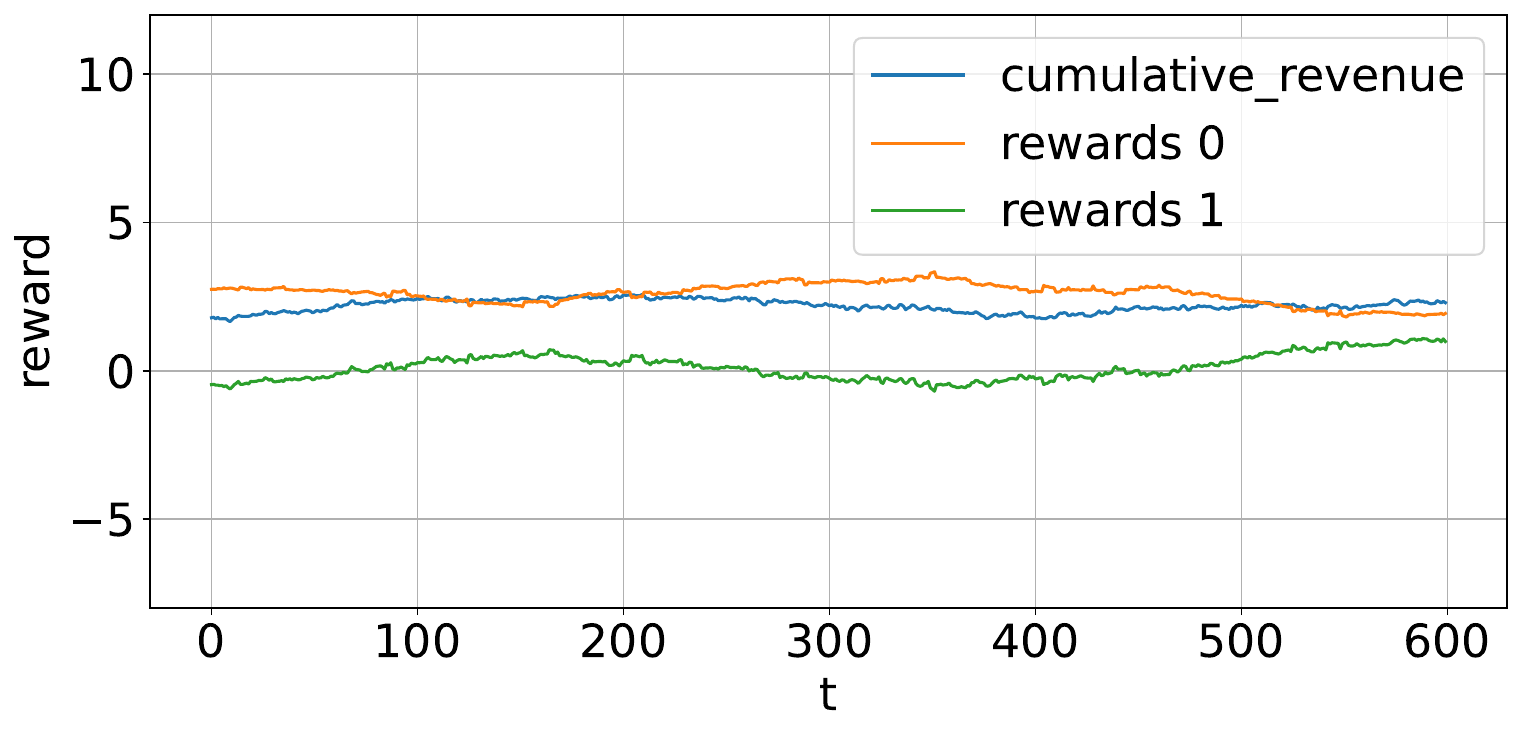}&
\includegraphics[width=0.32\textwidth]{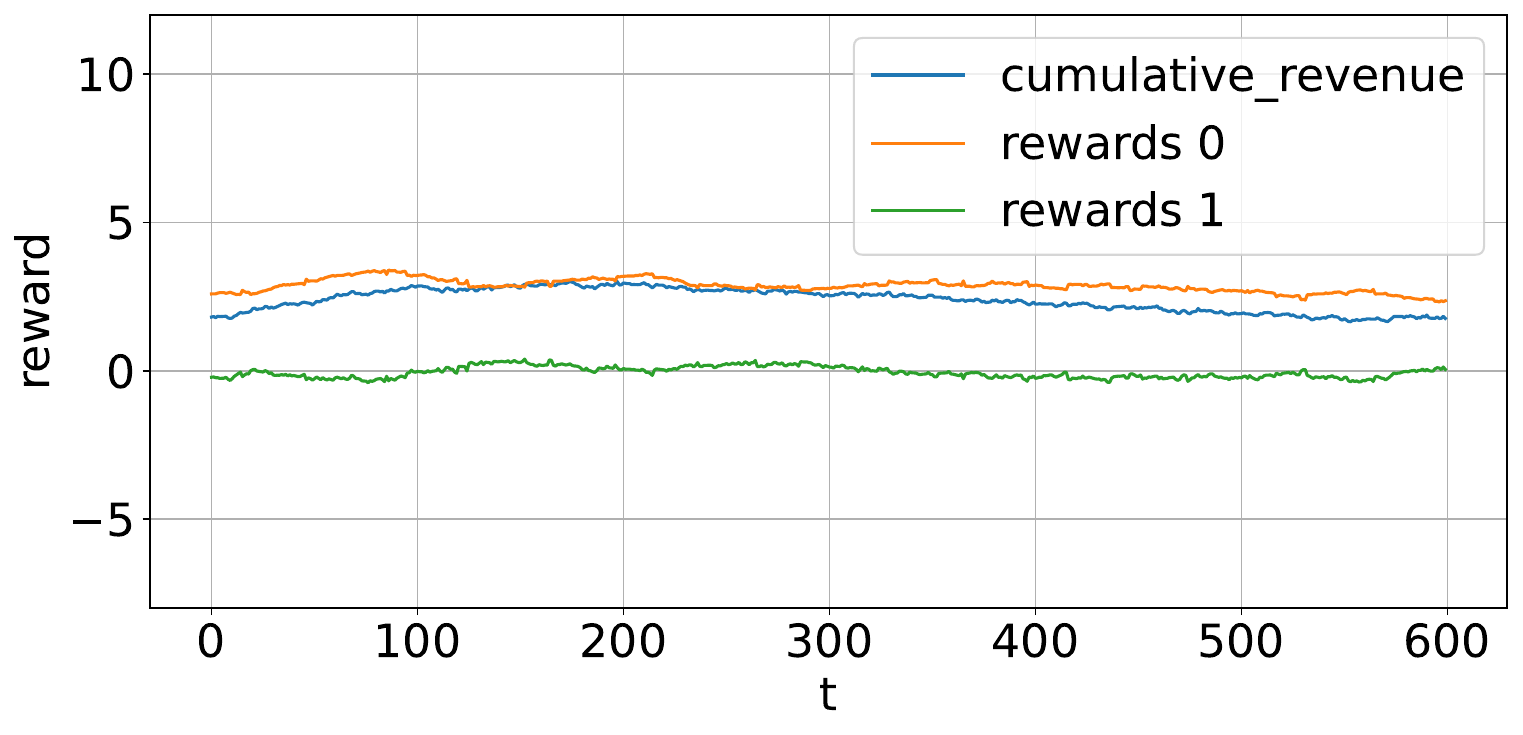}&
\includegraphics[width=0.32\textwidth]{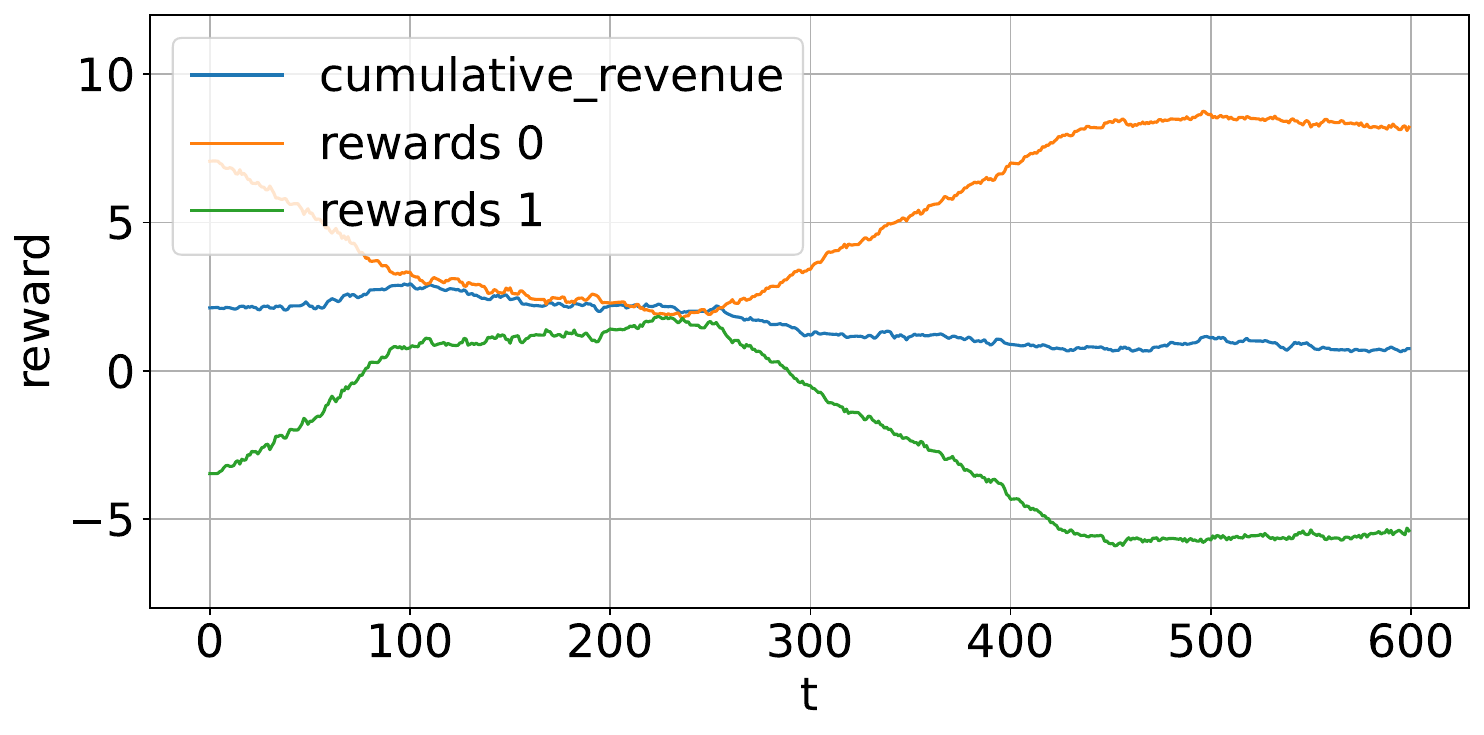}\\
\includegraphics[width=0.32\textwidth]{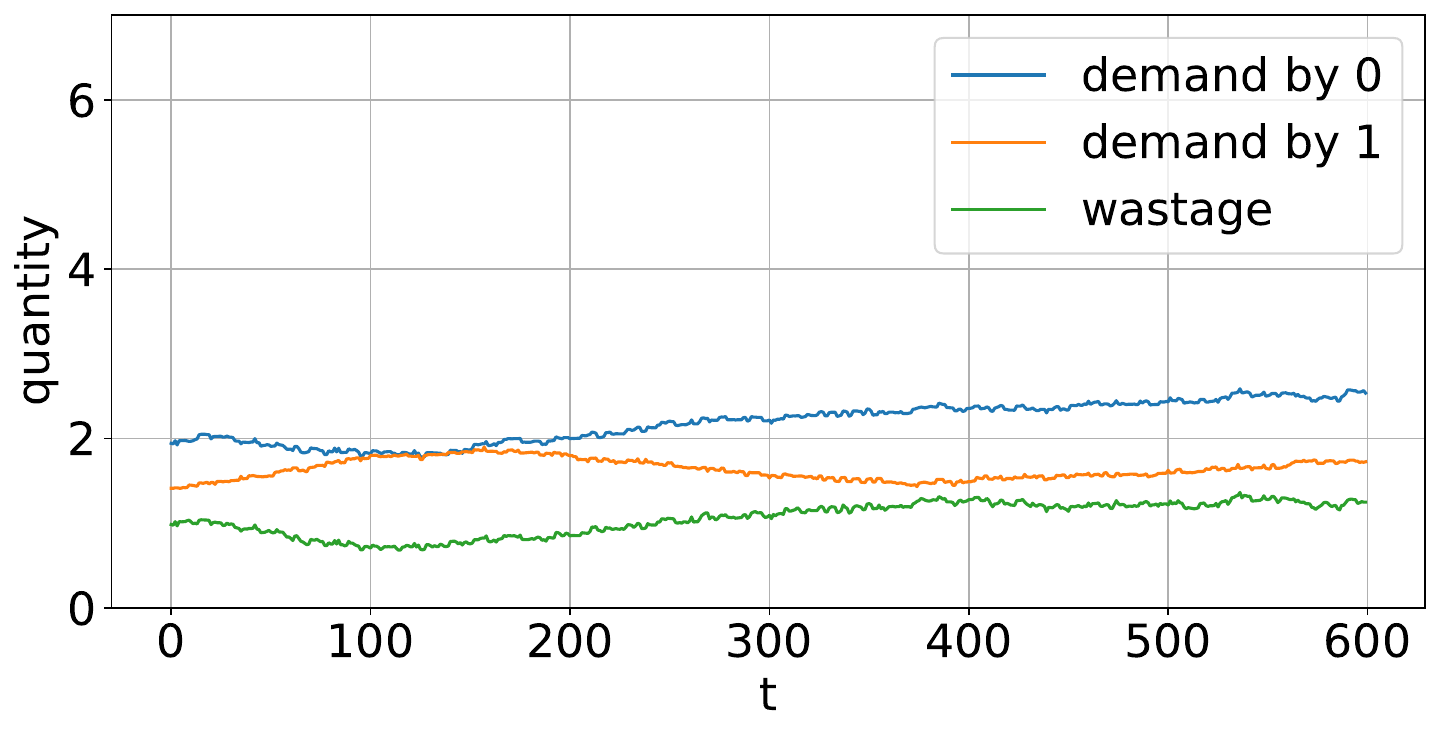}&
\includegraphics[width=0.32\textwidth]{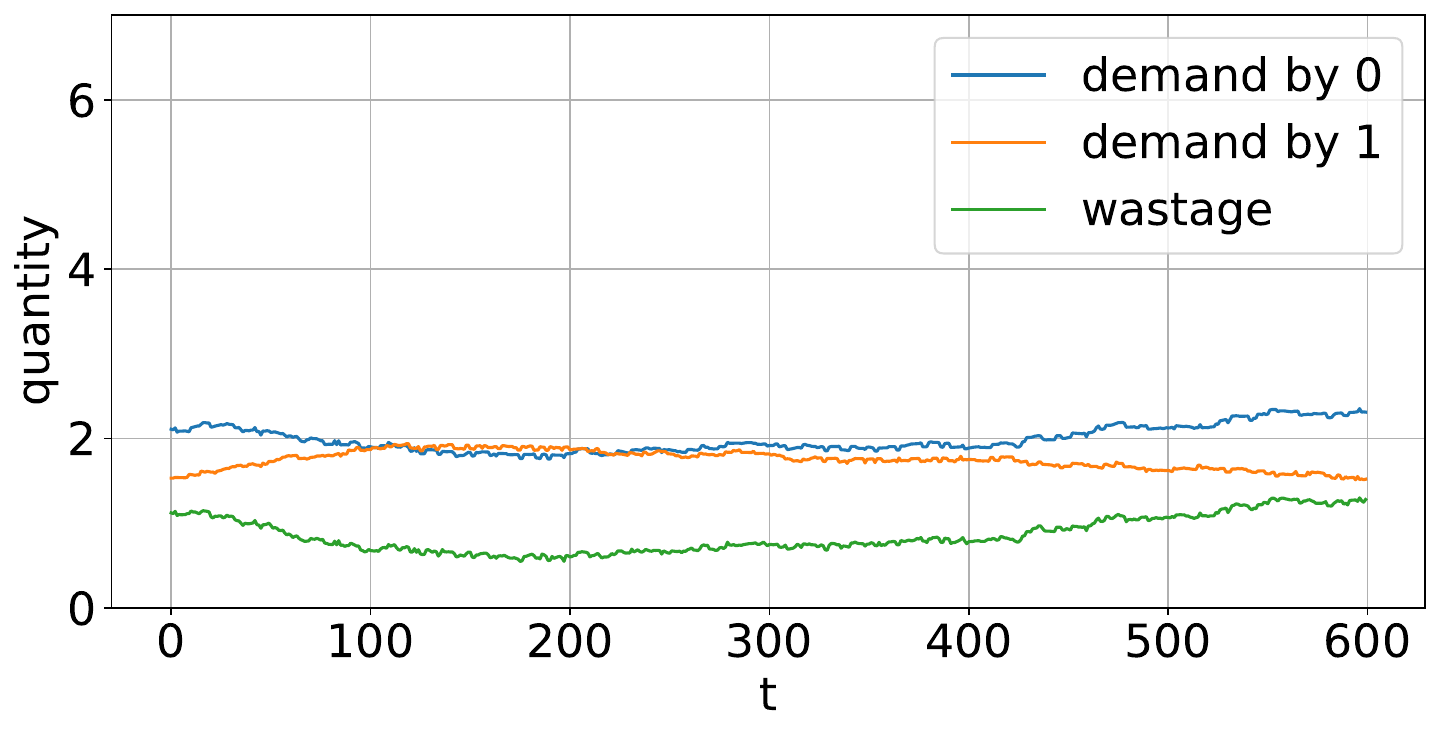}&
\includegraphics[width=0.32\textwidth]{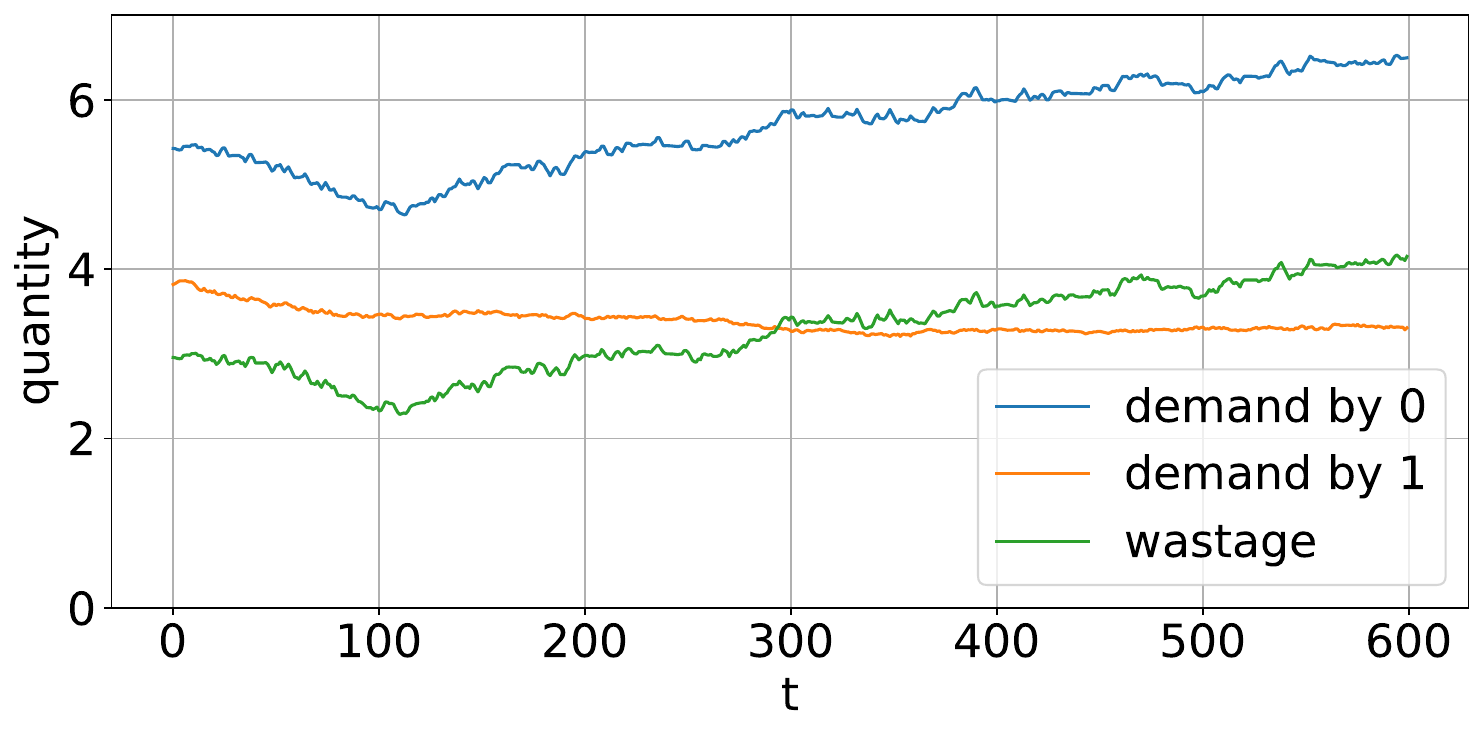}
\end{tabular}
\end{figure*}
Based on the setup in Section~\ref{sec:game}, we assume $z_0 = P(q_0)=0.5$, were $P(q_0)$ is the pre-defined raw material price; $D_1 = Q(p_1)=10-2p_1+0.05\epsilon$ where $Q(p_1)$ is the consumer demand and $\epsilon\in\mathcal{N}(0,1)$; we set $h_0=h_1=0.01$ and $w_0=w_1=0.1$. We share detailed configurations and experimental settings of actor and critic networks in supplementary material. In the 2PC scenario with privacy preserved data sharing (as shown in Figure~\ref{fig:Randomness1} and Table~\ref{tab:AvgRewardQuanityData}), the average rewards and costs are as follows: Reward0 is 2.496, Reward1 is 0.167, and the average wastage of the network is 1.085. The raw material cost is 0.5, and the average normalized cumulative profit (Total normalized revenue - operational cost - opportunity cost) stands at 2.1205. In No data share scenario: Avg reward0 is 6.129, avg reward1 = -2.933, avg wastage of the network is 3.411, raw mat cost is 0.5, normalized cumulative revenue is 1.4905. Therefore we observe that avg cumulative reward is 42.27\% better in 2PC than no data share. While, there is on average 68.19\% less wastage in 2 PC compared to no data share.

\begin{table}[t]
    \caption{Averages (across 800 iterations) of Reward and Quantity metrics in EDE and Secure 2PC modes keeping all external randomnesses the exact same (Randomness 1). R: reward, D: demand, W: wastage, subscript: Player Id} \label{tab:AvgRewardQuanityData}
    \begin{tabular}{|p{0.18\linewidth}|p{0.08\linewidth}|p{0.11\linewidth}|p{0.08\linewidth}|p{0.09\linewidth}|p{0.09\linewidth}|}
      \toprule 
      \bfseries  & \bfseries $R_0$ & \bfseries $R_1$ & \bfseries $D_0$ & \bfseries $D_1$ &  W\\      \midrule 
      EDE & 2.750 & $-0.07$ & 2.030 & 1.665 & 0.940\\
      \hline
      { Secure 2PC} & 2.496 & 0.167 & 2.207 & 1.604 & 1.085 \\
      \bottomrule 
    \end{tabular}
\end{table}

\begin{table}[t]
    \caption{Comparison of moving averages of rewards and demand quantities across the entire learning trajectory in Secure 2PC vs EDE mode. Metrics are same as table ~\ref{tab:AvgRewardQuanityData}} \label{tab:ComaparisonRewardQuanityData}
    \begin{tabular}{|p{0.12\linewidth}|p{0.11\linewidth}|p{0.11\linewidth}|p{0.11\linewidth}|p{0.11\linewidth}|}
      \toprule 
      \bfseries  & \bfseries $R_0$ & \bfseries $R_1$ & \bfseries $D_0$ & \bfseries $D_1$\\
      \midrule 
      MAE & 0.346 & 0.378 & 0.249 & 0.137\\
      \hline
      RMSE & 0.424 & 0.483 & 0.277 & 0.156 \\
      \bottomrule 
    \end{tabular}
\end{table}
Each agent undergoes an initial training phase using simulated data, without collaborating with the other agent. In practice, this simulation mimics market dynamics and is a feature in many enterprise supply chain management systems. The objective of this phase is to familiarize the agent with the MARL environment and to preliminarily train the actor-critic networks. This initial training allows the network to learn and explore without being directly exposed to the real-world, or the "go-live", environment. As illustrated in Figure~\ref{fig:Randomness1}, even before agents transition to using confidential data without explicit sharing, this preliminary training yields notable benefits, such as reduced wastage and increased cumulative revenue. For our study, we ran this simulated training for 9900 epochs. Once an agent has gained sufficient knowledge from this initial phase, it's ready to operate in the "go-live" environment, where all data is confidential. Here, the privacy-preserving mechanisms are essential. After the initial training, the "go-live" phase can continue indefinitely. In our tests, we operated in this environment for 20 epochs, striking a balance between comprehensive presentation and result sufficiency. For experimental purposes, the data in the "go-live" phase comes from a different simulation than the initial data, to emulate "out-of-distribution" samples. mirroring real-world scenarios.

\textbf{On Faithfulness:} To prove equivalence of secure computation and explicit exchange of private data we run the "go-live" simulation in 2 modes: {\em Explicit Data Exchange (EDE)} and {\em Secure 2PC}. All external randomnesses were kept the exact same in different modes of runs. The intention is to show that for a fixed set of external randomnesses (say {\em Randomness 1}) the results are equivalent. Table~\ref{tab:AvgRewardQuanityData} shows the closeness of various metric averages in 2 modes across all iterations with respect to Randomness 1. For a metric $M$, since individual values of $M(t)$ for $t\in[800]$, are subject to a lot of noise, pictorially we plotted the moving averages with a smoothing interval of 200. Figure~\ref{fig:Randomness1} shows these moving averages of various metrics. Figure~\ref{fig:Randomness1} also hints at the fact that, not only, the final average metric results are close in Secure 2PC and EDE, the learning trajectories are equivalent throughout the entire training time. Lemma~\ref{theorem} binds the deviations between floating point operations between secure and the non-secure runs. Table~\ref{tab:ComaparisonRewardQuanityData} shows a comparison of $M^\prime(t)$ values for different metrics obtained from 2 different modes. Table~\ref{tab:WeightsAndBiasDataPlayers} shows the MAE and RMSE estimates of the final weights and biases of various NN's obtained via 2 different modes: Secure 2PC and EDE. We report more experimental results in supplementary material.
\begin{table}[t]
\caption{Comparison of the final weights and biases of the neural networks in Secure 2PC mode vs EDE mode} \label{tab:WeightsAndBiasDataPlayers}
    \begin{tabular}{|l|c|c|}
      \toprule 
      \bfseries  & MAE & RMSE \\
      \midrule 
     Actor Player 0 & $2.37\times 10^{-3}$ & $3.84\times 10^{-3}$ \\
     \hline
     Critic Player 0 & $7.21\times 10^{-3}$ & $1.66\times 10^{-2}$ \\
     \hline
     Actor Target Player 0 & $9.69\times 10^{-4}$ & $1.56\times 10^{-3}$ \\
     \hline
     Critic Target Player 0 & $3.11\times 10^{-3}$ & $7.34\times 10^{-3}$ \\
     \hline
     Actor Player 1 & $2.69\times 10^{-3}$ & $4.24\times 10^{-3}$ \\
     \hline
     Critic Player 1 & $6.01\times 10^{-3}$ & $1.67\times 10^{-2}$ \\
     \hline
     Actor Target Player 1 & $1.14\times 10^{-3}$ & $1.90\times 10^{-3}$ \\
     \hline
     Critic Target Player 1 & $2.61\times 10^{-3}$ & $7.59\times 10^{-3}$ \\
      \bottomrule 
    \end{tabular}
\end{table}

\begin{table}[t]
\caption{Time and Communication of secure neural network operations (1 pass)} \label{tab:TimeAndCommunication}
        \begin{tabular}{|p{0.30\linewidth}|p{0.07\linewidth}|p{0.13\linewidth}|p{0.12\linewidth}|}
      \toprule 
      \bfseries  & Batch Size & Time (sec) & {Comm (GB)} \\
      \midrule 
      Action prediction & 1 & 1.014 & 0.04 \\
      \hline
      Actor Grad Update & 128 & 79.812 & 17.87 \\
      \hline
      Critic Grad Update & 128 & 66.540 & 14.43 \\
      \bottomrule 
    \end{tabular}
\end{table}

\textbf{Limitations and Future Work:} The secure privacy-preserving version of computation comes up with significant overhead in terms of time and communication due to cryptographic overheads. Table~\ref{tab:TimeAndCommunication} shows the time and communication of the frequently occurring NN operations in the secure 2PC mode. One complete iteration of training took 545.73 seconds in Secure 2PC and 0.035 seconds in EDE. Now a typical MARL set up should be run a significantly high number of epochs to obtain a reasonable equilibrium. If all the epochs are run with secure 2 PC in a brute force way, there will be huge computation and communication overhead. To address this in our current set up we first let the agents run an initial simulation where simulated data (not private sensitive data) are shared in the clear between them. This simulation step helps the agents to explore the game environment, gather enough experience and properly initialize their decision making neural networks. After that a relatively smaller number of epochs are run in secure 2 PC format with proper secret information. The high level API designs (F-\secfloat\ and B-\secfloat\ ) really come good here. This strategy only partly solves this issue. Improving over time and communication complexities of secure operations and lowering the deviation magnitudes of secure and non-secure versions further can open up more applications.

\section{Generalization and Conclusion}\label{sec:Generalizations}
This paper presents a secure multi-party computation (MPC) framework using floating-point operations for policy gradient approaches, specifically MADDPG. Specifically, the privacy preserving modules for MADDPG framework presented in this paper has demonstrated its effectiveness in a 2-party supply chain game. This paper empirically shows the integrity of such a solution. By significantly reducing wastage and improving revenue in industries involving multi-party decision making, these findings could lead to more efficient and secure operations, thereby benefiting the economy and society as a whole. As such, this framework enables previously unavailable data sharing solutions, marking a significant contribution to both the field of privacy-preserving machine learning and application research related to industrial operations.

It is important to note that its applicability extends to other 2-party MARL scenario that can be efficiently modeled in a similar game-theoretic fashion. In scenarios involving more than two parties, our framework can still be directly applied if all interactions among the parties can be reduced to a series of 2-party interactions. This is often the case in real-world supply chain scenarios, where most interactions occur pairwise and can therefore be modeled as a series of 2-party setups. If more than 2-party interaction is unavoidable, a non-colluding universal client-server model could be adopted. In this model, all parties secret share their private data between a client and a server, who then follow the 2-party privacy preserving algorithm. One may argue that if a trustworthy universal server is available then why can't all send their data in the clear to that universal server, get the computation done by the server and obtain the results? This will rule out the need of MPC. In the client-server setup, neither the client nor the server can extract any clear information as they only receive a share of every data. This is a less strict and more realistic assumption compared to the universal trustworthy server model. It's worth noting that this model contrasts with frameworks like federated learning, which require a universal trustworthy server. In this model, the server cannot see all data in the clear, thus providing an additional layer of privacy. Beyond its direct application in supply chain scenarios, the proposed framework also holds potential for broader industrial applications, such as data center task management \citep{Nishtala2020TwigMT}, energy grid supply \citep{Zhang2022MultistepMR}, communication networks \citep{Zhou2021IntelligentAC}, media and entertainment \citep{Jin2020JointQC}, among others.
\clearpage
\newpage

\bibliographystyle{ACM-Reference-Format} 
\bibliography{references}

\pagebreak
\title{Supplemental Materials}
\maketitle

\section{Organization of the Paper}
The cryptographic background of MPC and
{\secfloat} are described in Section 2. Section 3 builds a set of high-level privacy-preserving secure neural network modules, a forward-pass gadget (F-\secfloat\ API) and a backpropagation gadget (B-\secfloat\ API). These modules (APIs) secure the computation of forward and backward passes. Section 4 sets up the two-player supply chain model as described in \citet{li2022general}.
Section 4 also describes the game setup between the 2 supply chain players sharing data within a MARL framework. Here each player is optimizing towards their own objectives while responding strategically to other player's actions using MADDPG. This section illustrates how this is achieved in a secure-privacy preserving way by invoking suitable {\secfloat}-based neural network modules as developed in Section 3. In section 5, we present results for secure computation on the private data of 2 parties using our framework. Comparisons are made against the baseline case where the players explicitly exchange their private data, demonstrating the integrity of our solution that achieves equivalent results. The increased computation time and communication complexity associated with secure computation are also discussed.
Section 6 discusses the generalization of the 2PC setup to an MPC setup and also beyond supply chains. 
\section{Cryptographic Primitives - Addendum}
\subsection{A bird's eye view of an MPC protocol}\label{MultiplicationGate} We shall first introduce the notion of secret sharing and then show how to compute any arithmetic circuit securely, for more details see \citep{10.1145/62212.62213}.

\noindent {\bf Secret sharing.} A $(t+1)$-out-of-$n$ secret sharing scheme is a scheme via which a dealer shares a secret $s$ amongst $n$ parties in such a way that any subset of $(t+1)$ or more parties can reconstruct the secret uniquely by combining their shares but any subset of $t$ or fewer parties learn nothing about the secret. Shamir's secret sharing scheme \citep{Shamir1979HowTS} uses the fact that in a field (say $\mathbb{Z}_p$, for a prime $p>n$), given $(t+1)$ points on the 2-D plane $(x_1,y_1),...,(x_{t+1},y_{t+1})$ with unique $x_i$'s, there exists a unique polynomial $q(x)$ of degree at most $t$ such that $q(x_i)=y_i$ for all $i$ and $q(x)$ can be efficiently reconstructed by interpolation using the Lagrange basis polynomials. So, in order to share a secret $s$ among $n$ parties in the $(t+1)$-out-of-$n$ mode, a party first chooses a random polynomial $q(x)$ with a degree at most $t$ with the constraint $q(0)=s$. So she basically randomly chooses $a_1,...,a_t\in \mathbb{Z}_p$, sets $a_0=s$ and $q(x)={\sum}_{i=0}^ta_i.x^i$. Then, she sends the $i$'th party $y_i=q(i)$, the $i$'th share of $s$. Clearly, $(t+1)$ or more number of parties can reconstruct $q(x)$ and thus $q(0)=s$ by interpolation of their shares but for $t$ or less number of parties $q(x)$ remains random and therefore $s=q(0)$ as well.

\noindent{\bf Computing an arithmetic circuit.} Now, let $n$ parties hold the secret shares of 2 values $s_a$ and $s_b$ via 2 polynomials $a(x)$ and $b(x)$ respectively, i.e., $a(0)=s_a$, $b(0)=s_b$ and the shares of $s_a$ and $s_b$ held by the $i$'th party are $a(i)$ and $b(i)$ respectively. We outline how they can compute the secret shares of $(s_a+s_b)$ and $s_a.s_b$. Inductively, they can then use the same procedure to compute any arithmetic function securely.

\textbf{For the addition gate}, the $i$'th party can just locally compute $a(i)+b(i)$, her share for $(s_a+s_b)$. Since, $c(x)=a(x)+b(x)$ is again a polynomial with degree at most $t$ and $c(0)=a(0)+b(0)=s_a+s_b$,  $c(i)=a(i)+b(i)$ for $i\in[n]$ give a valid $(t+1)$-out-of-$n$ secret share for $c(0)$.

\textbf{For the multiplication gate} $s_a.s_b$, for $i\in[n]$,   $a(i).b(i)$  are secret shares in a $(2t+1)$-out-of-$n$ mode (not in a desired $(t+1)$-out-of-$n$ mode) since $c(x)=a(x).b(x)$ is now a polynomial with degree at max $2t$. A reduction step is required.

For degree reduction from $2t$ to $t$, the parties need to hold 2 independent secret sharing of an unknown random value $r$, the first via a degree $2t$ polynomial $R_{2t}(x)$ and the second via a degree $t$ polynomial $R_t(x)$ with $R_{2t}(0)=R_t(0)=r$. They can then locally compute the shares of the degree $2t$ polynomial $d(x)=c(x)-R_{2t}(x)$ and can reconstruct $d(0)=c(0)-r$ by combining all $d(i)$'s for $i\in [n]$. Here, $t<\frac{n}{2}$ is required to ensure that all the $n$ parties collectively can reconstruct the $2t$ degree polynomial. Now since, $c'(x)=R_t(x)+d(0)$ is a degree $t$ polynomial and $c'(0)=r+c(0)-r=c(0)=a(0).b(0)$, $c^\prime(i)$'s for $i\in [n]$ give a valid $(t+1)$-out-of-$n$ secret sharing of $a(0).b(0)$, as desired. Now, in order to achieve the 2 independent secret sharing of a random unknown value r, for all $i\in[n]$, the $i$'th party shares a random $r_i$ via a degree $2t$ polynomial $R_{2t}^i(x)$ and a degree $t$ polynomial $R_{t}^i(x)$. Then, $r=\sum_{j=1}^nr_j$ is unknown to all $i$ and $R_{2t}(i)=\sum_{j=1}^nR_{2t}^j(i)$ and $R_{t}(i)=\sum_{j=1}^nR_{t}^j(i)$ become the secret shares of $r$ for the degree $2t$ polynomial $R_{2t}(x)$ and the degree $t$ polynomial $R_{t}(x)$ respectively.

\noindent{\bf MPC sub-protocol invariants.} An important feature of most MPC protocols is that they provide a way for parties to begin with secret shares of values $x_1, ..., x_n$ and execute a protocol such that they end up with secret shares of $f(x_1, x_2, \cdots, x_n)$. This invariant is maintained for several simple functions. Once this is available for a set of functions $f_1, \cdots, f_k$, they can be composed to compute the composition of these functions. In the above example, $f_1$ and $f_2$ were addition and multiplication respectively. 
\section{High Level APIs}
\textbf{High-Level APIs: }\secfloat\ guarantees the security of elementary math operations - such as comparison, addition, multiplication, division, and functions of floating-point values - viz sine, tangent, logarithm, exponent, etc. Additionally it guarantees the security of arbitrary compositions of these operations and function. We then compose them in a hierarchical way to build secure high-level functionalities - matrix/vector multiplications, matrix/vector additions, various activation function calculations (for eg., relu, sigmoid, tanh, etc.), and exponentiation. Using this we then develop the high level secure APIs: For forward pass: $\forward(\Tilde{X_0},\Tilde{X_1},\mathcal{W})$; For $\nabla_{\mathcal{W}}N$: $\backward_{\mathcal{W}}(\Tilde{X_0},\Tilde{X_1},\mathcal{W})$; For $\nabla_{\Tilde{X_i}}N$: $\backward_{\Tilde{X_i}}(\Tilde{X_0},\Tilde{X_1},\mathcal{W})$; For $\nabla_{\mathcal{W}}L(N,t)$: $\backwardl_{\mathcal{W}}(\Tilde{X_0},\Tilde{X_1},\mathcal{W},t)$, where $t$ is the target.

\section{Reward Structures for Individual Players}
\begin{figure*}[htbp]
    \centering
    \begin{tikzpicture}[xscale=1,yscale=1]
        \node (A) [draw,circle,minimum size=1cm,inner sep=2pt, align=center, fill=white!80!blue] at (-4.5,0) {Raw\\ Materials\\ Market};
        \node (B) [draw,circle,minimum size=1cm,inner sep=2pt, fill=white!80!purple] at (-1.5,0) {Player 0};
        \node (C) [draw,circle,minimum size=1cm,inner sep=2pt, fill=white!80!purple] at (1.5,0) {Player 1};
        \node (D) [draw,circle,minimum size=1cm,inner sep=2pt, align=center, fill=white!80!blue] at (4.5,0) {Consumer\\ Market};
        \draw[thick,-stealth,color=black!30!violet] (A.30) -- (B.137) node[midway,above] {$P(q_0)$};
        \draw[thick,-stealth,color=black!30!orange] (B.180) -- (A.0) node[near start,below] {$q_0$};
        \draw[thick,-stealth,color=black!50!green] (A.330) -- (B.223) node[near start,below] {$q_0$};
        \draw[thick,-stealth,color=black!30!violet] (B.35) -- (C.145) node[near start,above] {$p_0$};
        \draw[thick,-stealth,color=black!30!orange] (C.180) -- (B.0) node[near start,below] {$q_1$};
        \draw[thick,-stealth,color=black!50!green] (B.325) -- (C.215) node[near start,below] {$d_{10}$};
        \draw[thick,-stealth,color=black!30!violet] (C.43) -- (D.148) node[near start,above] {$p_1$};
        \draw[thick,-stealth,color=black!30!orange] (D.176.5) -- (C.5) node[midway,below] {$Q(p_1)$};
        \draw[thick,-stealth,color=black!50!green] (C.317) -- (D.213) node[near start,below] {$d_{c1}$};

        \path ([xshift=1cm,yshift=0.5cm]current bounding box.north east)
 node[matrix,anchor=north west,cells={nodes={font=\sffamily,anchor=west}},
 draw,thick,inner sep=1ex]{
  \draw[-latex,color=black!30!violet](0,0) -- ++ (0.6,0); & \node{\small{Price Charged}};\\
  \draw[-latex,color=black!30!orange](0,0) -- ++ (0.6,0); & \node{\small{Quantity Demanded}};\\
  \draw[-latex,color=black!50!green](0,0) -- ++ (0.6,0); & \node{\small{Realized Demand}};\\
 };
    \end{tikzpicture}
    \caption{Two Player Single Commodity Supply Chain}\label{fig:SupplyGame}
\end{figure*}
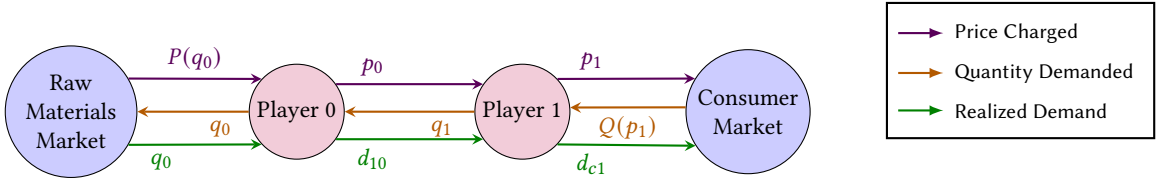

At a given time stamp $t$, Fig~\ref{fig:SupplyGame} depicts the price, quantity and demand dynamics of the game. We assume the consumer market demand to be $Q(p_1)$, a function of price charged, and the raw material price to be $P(q_0)$, a function of quantity demanded. Moreover, $d_{10}=min\left(q_1,x_0\right)$  and $d_{c1}=min\left(Q(p_1),x_1\right)$ where $x_i$ is the stock level of Player $i$ at that time. Then reward expressions for Player 0 and 1 become:
\begin{equation}\label{eqn:RewardEqnPlayer0}
    r_0 = p_0d_{10} - P(q_0)q_0 - h_0\left(x_0-d_{10}\right) - w_0\left(q1-d_{10}\right)
\end{equation}
\begin{equation}\label{eqn:RewardEqnPlayer1}
    r_1 = p_1d_{c1} - p_0d_{10} - h_1\left(x_1-d_{c1}\right) - w_1\left(Q(p_1)-d_{c1}\right)
\end{equation}

\section{Secure MADDPG: Details}
Let's consider the action inference step of Player $i$ given by $a_i=\pi_i(s_i,s_{1-i}\vert\theta_i)$. Here $s_i$ and $s_{1-i}$ are held by Player $i$ and $1-i$ respectively and don't share it in the clear. Player $i$ also holds the weights and biases of the NN given by $\theta_i$. First they pre-process their data according pre-processing step in Section 3 to make the data compatible for the \forward\ API call. Essentially, Player 0 prepares $\tilde{s_0}=s_0\:||\:0^d$ and Player 1 prepares $\tilde{s_1}=0^d\:||\:s_1$ where $s_0,s_1\in\mathbb{R}^d$. Now, updated version of the action inference equation becomes $a_i=\forward(\tilde{s_i}, \widetilde{s_{1-i}},\theta_i)$. We describe how the MADDPG gradient expressions for Player $i$ transform in this secure hybrid API integrated framework. In the format we propose, which involves explicit data exchange, the process happens as follows: Player $i$ samples a minibatch $B_i\subset \mathcal{D}_i$ from her local replay buffer. After this, she exchanges it with Player $-i$. It's important to note that both minibatches have the same sampling randomness. In no explicit data exchange realm, first Player $i$ samples an index set $\mathcal{I}_i\subset |\mathcal{D}_i|$ and sends that explicitly to Player $(1-i)$. This is essentially the set of indices that they sample for their minibatches. So, for both $j\in\{0,1\}$, $B_j\gets \{\bigl(s_j,a_j,r_j,s^\prime_j\bigr)_k\in \mathcal{D}_j\}$ where $k\in\mathcal{I}_i$. Then they row wise stack up $s_j,a_j,r_j,s^\prime_j$ vectors to obtain $S_j,A_j,R_j,S^\prime_j$ matrices. The appended state-action matrix $V_j=S_j][A_j$. They pre-process these matrices, note pre-processed matrices have overhead tilde sign. Now, $j\in\{0,1\}$ both predict their next actions by $A^\prime_j=\forward(\Tilde{S^\prime_j},\widetilde{S^\prime_{1-j}},\theta^\prime_j)$, followed by getting the appended next state, next action matrix $V_j^\prime=S_j^\prime][A_j^\prime$ and pre-process that to obtain $\tilde{V_j^\prime}$.

\section{Time Complexity Theorems Derivations}
\begin{theorem}\label{Preprocessing-theorem-maddpg}
    Computational complexity for the entire lifetime of the 2 party secure MADDPG using F-\secfloat\ and B-\secfloat\ gadgets is $\mathcal{O}(f (\vert S\vert + \vert A \vert) n)$ where $\vert S\vert$ represents the state-space size, $\vert A\vert$ the action space size, and 'n' the total number of iterations (Number of epochs $\times$ Iterations per epoch). 
\end{theorem}
\textbf{Proof:} For the purpose of derivation we consider 2 key categories of computation for each iteration: a) Computation of the forward, backward pass; b) Pre-processing computation. Let number of computations for one pass of clear-text, without any cryptography computation, of the MADDPG be $f$. During pre-processing, the party has to sit idle for that entire time as no other computation can happen. Using high level gadgets, pre-processing is required for both the actor \& critic networks. In case of critic network as shown in lines 24 and 25 of Algorithm 1 (main paper), there are 1 pre-processing steps on states ($\Tilde{S^\prime_j}$ line 16) and 2 pre-processing steps on state, action pairs ($\Tilde{V_j}$ in line 16 and $\Tilde{V^\prime_j}$ in line 21). Then, for a single iteration the number of computations will be $
c_{critic} = f \times (3\vert S\vert + 2\vert A\vert)$.
In case of actor network (lines 27 and 28 of Algorithm 1 in main paper), there is one additional state pre-processing required on states ($\Tilde{S_j}$ in line 16). So, the computations will be $c_{actor} = f \times \vert S\vert$. The total computations are $c = f \times (3\vert S\vert + 2\vert A\vert) + f \times \vert S\vert$. Therefore, complexity for all the iteration becomes $\mathcal{O}(f (\vert S\vert + \vert A \vert) n)$, where $n$ is the number of iterations.
\begin{theorem}\label{Preprocessing-theorem-native}
Computational complexity for the entire lifetime of the 2 party MADDPG using native \secfloat\ implementation is  $\mathcal{O}(fln)$ in the best case scenario and $\mathcal{O}(f^2 (\vert S\vert + \vert A\vert )n)$ in the worst case scenario, where $\vert S\vert$ represents the state-space size, $\vert A\vert$ the action space size, $l$ is the number of actor-network nodes, and 'n' the total number of iterations (Number of epochs $\times$ Iterations per epoch). 
\end{theorem}
\textbf{Proof:} For this derivation, we follow a similar scheme as in the proof of Theorem~\ref{Preprocessing-theorem-maddpg}. The key change in the native implementation is that in addition to all the pre-processing steps, least the other party also needs to share the weights of the entire actor network. This means that, without the high level gadgets, an additional pre-processing step would need to be included for all the weights of the actor network, as shown in Figure 3 (main paper). We have $c_{critic} = f \times (3\vert S\vert + 2\vert A\vert))$. In the best case scenario, where Party 2 already has computed the actor network weights when Party 1 requires it, then $c_{actor} = f \times (\vert S\vert + l)$, where $l$ is the total number of nodes in the actor network. Total computations become $c =  f \times (\vert S\vert + l) + f \times (3\vert S\vert + 2\vert A\vert))$. Since $l \gg \vert S\vert + \vert A\vert$ the computational complexity becomes $\mathcal{O}(fln)$,
In the worst case, in order for the second party to share all the (cryptographic) weights of the actor network with the first party, the first party will have to wait for the second party to finish computing all the weights. Assuming both parties have the same (or similar) number of computations. The computations for actor network in this case will be:
\begin{align*}
\text{Party 1 } c_{actor} &= f \times (\vert S\vert + l + \text{Party 2 } (c_{critic} + c_{actor}))\\
&= f \times(\vert S\vert + l + f \times (3\vert S\vert + 2\vert A\vert) + f \times \vert S\vert + \text{Party 1 } c^{\{1\}}),
\end{align*}
where $c^{\{1\}}$ is the total computations for Party 1 from the previous time step. Assuming that $c^{\{1\}}$ is already calculated and stored in the previous time step, and that lookup computations are much smaller than the multiplication \& addition computations. Therefore the total computations become,
\begin{align*}
c &= f \times (3\vert S\vert + 2\vert A\vert) + f \times(\vert S\vert + l + f \times (3\vert S\vert + 2\vert A\vert) + f \times \vert S\vert)\\
&=f \times (3\vert S\vert + 2\vert A\vert) + f \times\vert S\vert + fl+ f^2 \times (3\vert S\vert + 2\vert A\vert) + f^2 \times \vert S\vert.
\end{align*}
Now $f \gg \vert S \vert + \vert A \vert$, therefore the computational complexity becomes $\mathcal{O}(f^2 (\vert S\vert + \vert A\vert )n)$, where $n$ is the number of iterations.
\section{Actor Critic Network Configurations, Hyper-parameters and Experimental Settings}
Each party is hosted via a 16-core, 2.4 GHz VM CPU with 126 GB RAM, with the communication speed between the machines being 15.2 Gbps. 
With this setting, we first ran a 9900 epoch (with each epoch being a 40-step trajectory) simulation of 2 parties where only simulated data (and no secret data) were shared between the parties. The {\em external randomnesses} are random state initialization at the start of each epoch, random noises added to the actions, and random sampling of mini-batches. This initial 9900 epoch simulation was an "exploration" phase where the parties collected a lot of transition and reward history in their replay buffers and set up their neural networks. Table~\ref{tab:ActorCriticNetworks} shows configuration details, batch size, some other hyper-parameters of the Actor and Critic neural networks used in the 2 party supply game experiment set up described in section 5 of the main paper.
\begin{table}[htp]
    \centering
    \small
    \begin{tabular}{l|c|c|}
      \toprule 
      ~ & Actor & Critic \\
      \midrule 
      No of Linear Layers & 3 & 3 \\
      \hline
      Input Dimension & 10 & 14\\
      \hline
      Hidden Dimension & 128 & 128 \\
      \hline
      Output Dimension & 2 & 1\\
      \hline
      Activation Functions & ReLU, Sigmoid & ReLU, Identity\\
      \hline
      Batch size for Back-prop & 128 & 128 \\
      \hline
      Learning rate & $10^{-4}$ & $10^{-3}$\\
      \hline
      Optimizer & Adam & Adam \\
      \bottomrule 
    \end{tabular}
    \caption{Actor Critic Network Configurations and Hyper-parameters} \label{tab:ActorCriticNetworks}
\end{table}
\section{Additional Results for a Different Set of fixed External Randomnesses (Randomness 2)}
\begin{figure*}
   \centering
\begin{tabular}{ccc}
Secure 2PC&Explicit Data Exchange&No Data Sharing\\
\includegraphics[width=0.3\textwidth]{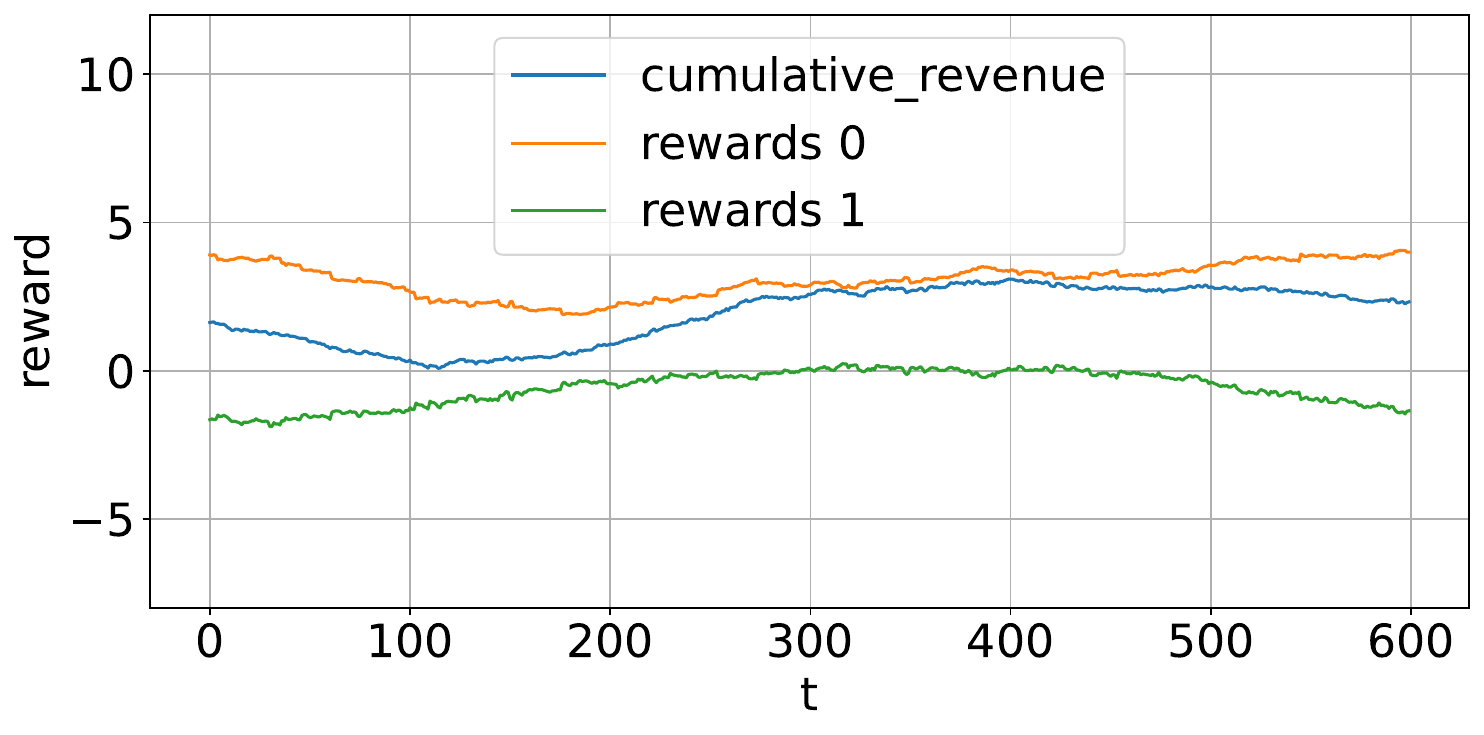}&
\includegraphics[width=0.3\textwidth]{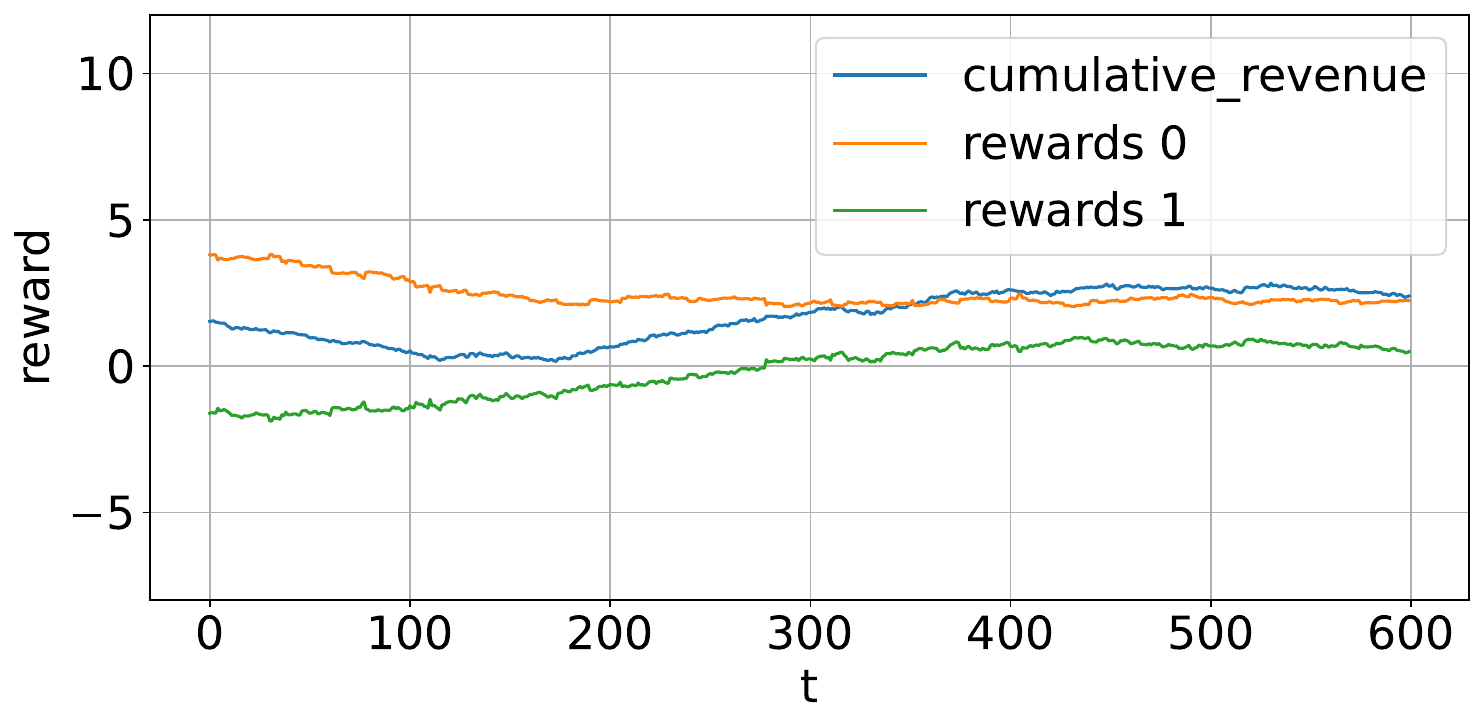}&
\includegraphics[width=0.3\textwidth]{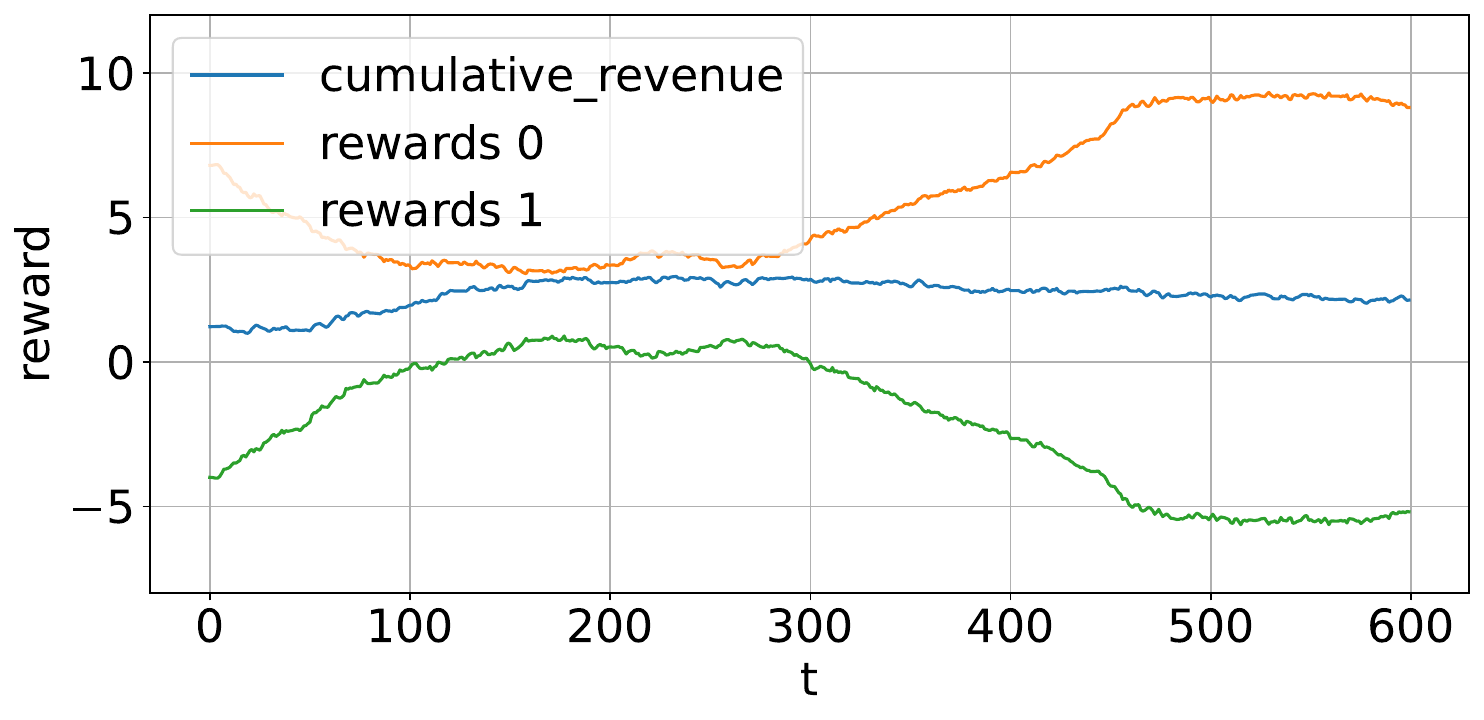}\\
\includegraphics[width=0.3\textwidth]{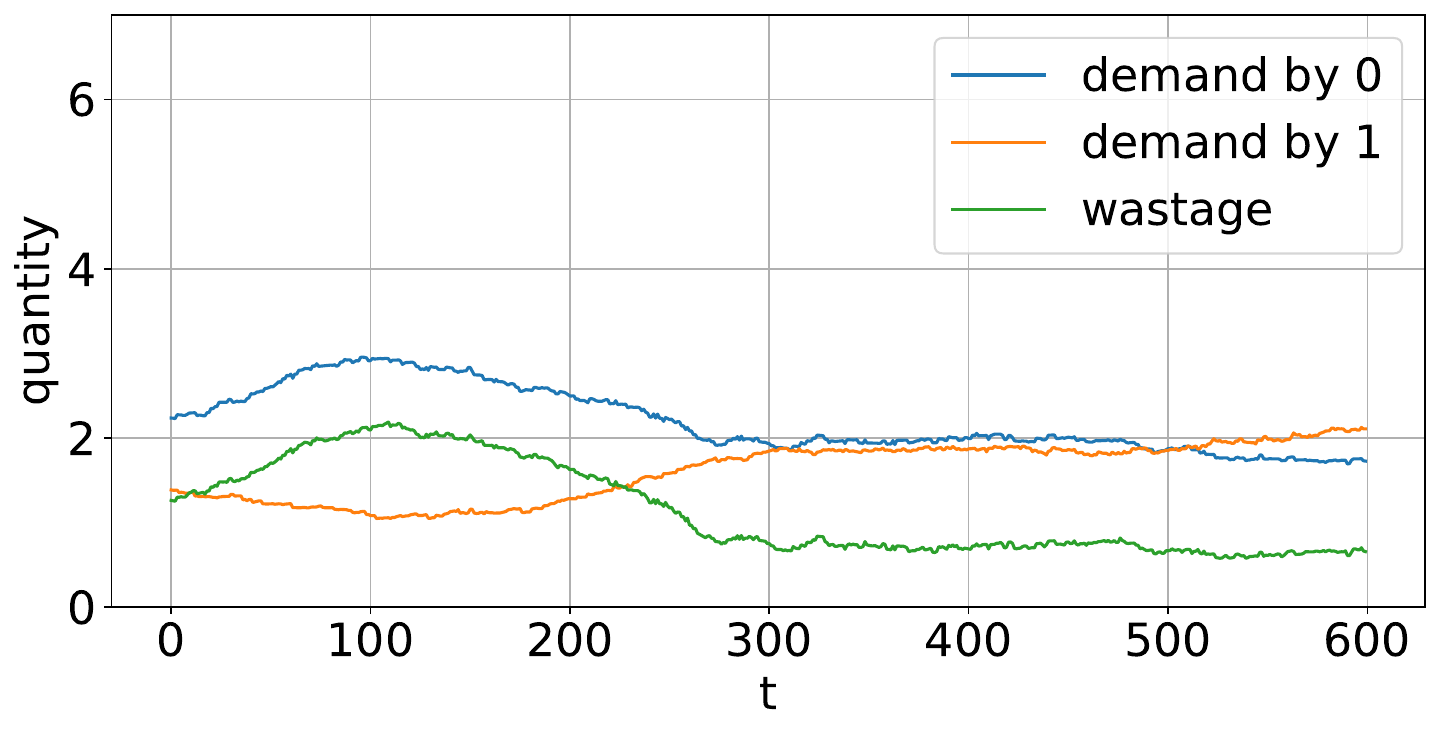}&
\includegraphics[width=0.3\textwidth]{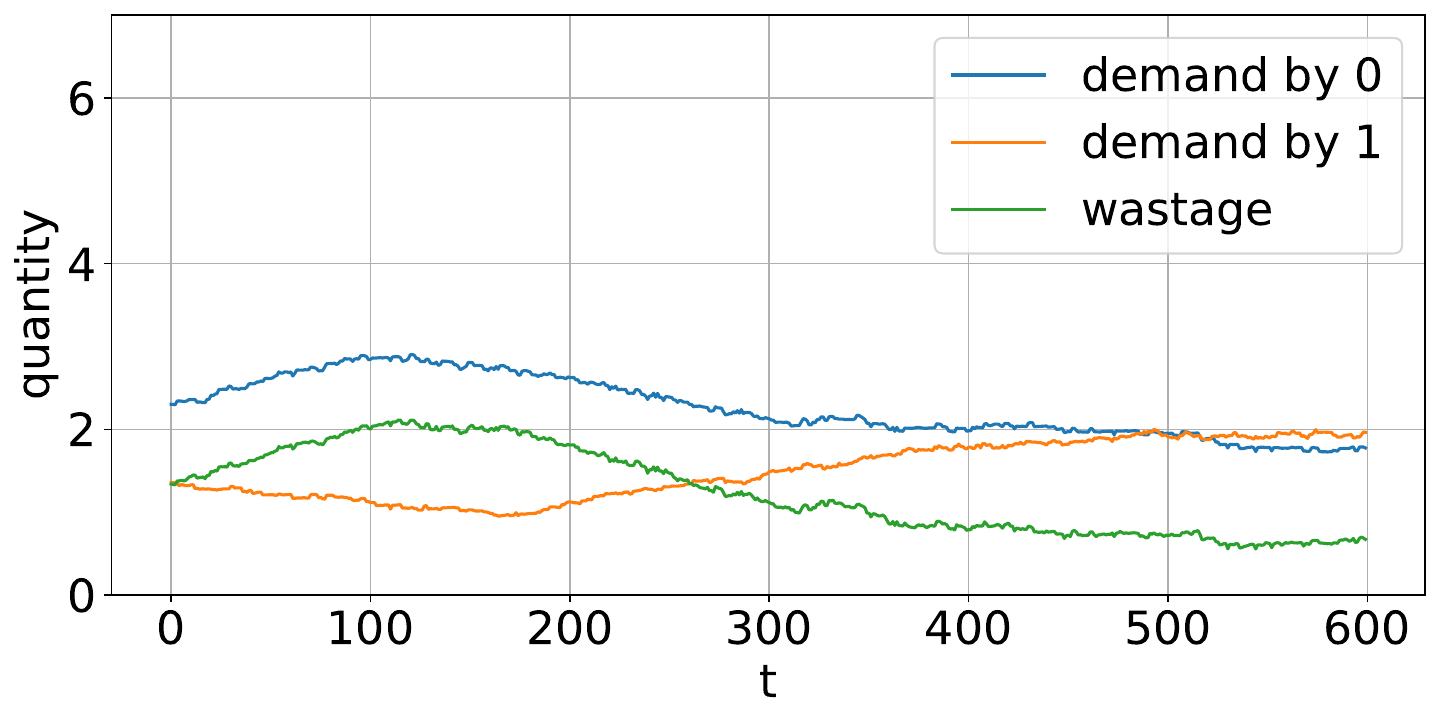}&
\includegraphics[width=0.3\textwidth]{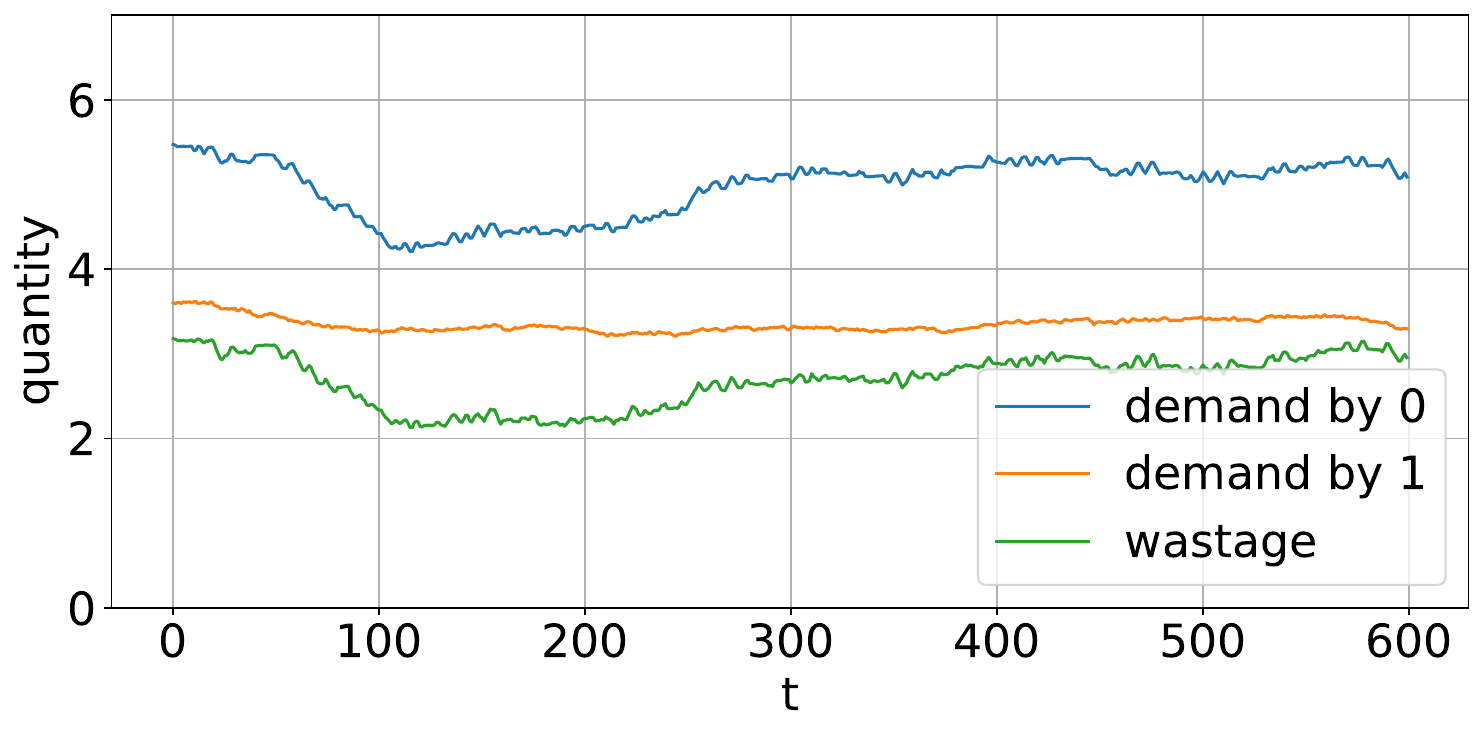}
\end{tabular}

    \caption{Reward and Quantity plots in different modes of execution keeping all external randomnesses the exact same across different modes (Fixed Randomness 2)}
    \label{fig:Randomness2} 
\end{figure*}

In section 5.1 of the main paper, we provided plots and statistics to show that secure computation on private data yields equivalent results to the case of explicit data exchange in terms of the metrics of supply game when all external randomnesses are kept fixed. We showed plots and statistics for a fixed set of external randomnesses viz, Randomness 1. Here we give the same plots and statistics for a different set of all external randomnesses viz, Randomness 2. Figure~\ref{fig:Randomness2} shows the moving averages of various metrics in different modes of execution. Table~\ref{tab:AvgRewardQuanityData} shows the closeness of various metric averages across 800 iterations in 2 modes explicit data exchange (EDE) and secure 2PC across all iterations with respect to Randomness 2. Table~\ref{tab:ComaparisonRewardQuanityData} shows a comparison of moving averages of different metrics obtained from 2 different modes across the entire learning trajectory. Table~\ref{tab:WeightsAndBiasDataPlayers} shows the MAE and RMSE estimates of the final weights and biases of various NN's obtained via 2 different modes.

\begin{table}
    \centering
    \begin{tabular}{l|c|c|c|c|c|}
      \toprule 
      \bfseries  & \bfseries $R_0$ & \bfseries $R_1$ & \bfseries $D_0$ & \bfseries $D_1$ &  W\\      \midrule 
      EDE & 2.616 & -0.238 & 2.169 & 1.551 & 1.149\\
      \hline
      Secure 2PC & 3.344 & -0.828 & 2.114 & 1.654 & 1.057 \\
      \bottomrule 
    \end{tabular}
    \caption{Averages (across 800 iterations) of Reward and Quantity metrics in EDE and Secure 2PC modes keeping all external randomnesses the exact same (Randomness 2). R: reward, D: demand, W: wastage, subscript: Player Id} \label{tab:AvgRewardQuanityData}
\end{table}

\begin{table}
    \centering
    \small
    \begin{tabular}{l|c|c|c|c|}
      \toprule 
      \bfseries  & \bfseries $R_0$ & \bfseries $R_1$ & \bfseries $D_0$ & \bfseries $D_1$\\
      \midrule 
      MAE & 0.499 & 0.880 & 0.081 & 0.123\\
      \hline
      RMSE & 0.578 & 1.011 & 0.105 & 0.164\\
      \bottomrule 
    \end{tabular}
    \caption{Comparison of moving averages of rewards and demand quantities across the entire learning trajectory in Secure 2PC vs EDE mode for a fixed set of external randomnesses (Randomness 2). Metrics are same as table ~\ref{tab:AvgRewardQuanityData}} \label{tab:ComaparisonRewardQuanityData}
\end{table}

\begin{table}
    \centering
    \begin{tabular}{l|c|c|}
      \toprule 
      \bfseries  & MAE & RMSE \\
      \midrule 
     Actor Player 0 & $2.63\times 10^{-3}$ & $4.12\times 10^{-3}$ \\
     \hline
     Critic Player 0 & $7.84\times 10^{-3}$ & $1.75\times 10^{-2}$ \\
     \hline
     Actor Target Player 0 & $1.22\times 10^{-3}$ & $1.90\times 10^{-3}$ \\
     \hline
     Critic Target Player 0 & $3.41\times 10^{-3}$ & $7.79\times 10^{-3}$ \\
     \hline
     Actor Player 1 & $2.55\times 10^{-3}$ & $4.06\times 10^{-3}$ \\
     \hline
     Critic Player 1 & $5.69\times 10^{-3}$ & $1.47\times 10^{-2}$ \\
     \hline
     Actor Target Player 1 & $1.02\times 10^{-3}$ & $1.63\times 10^{-3}$ \\
     \hline
     Critic Target Player 1 & $2.31\times 10^{-3}$ & $6.08\times 10^{-3}$ \\
      \bottomrule 
    \end{tabular}
    \caption{Comparison of the final weights and biases of the neural networks in Secure 2PC mode vs EDE mode for a fixed set of external randomnesses (Randomness2)} \label{tab:WeightsAndBiasDataPlayers}
\end{table}
\section{Additional Results for a Different Price-Demand Distribution}
\begin{figure*}
   \centering
\begin{tabular}{cc}
Explicit Data Exchange&No Data Sharing\\
\includegraphics[width=0.4\textwidth]{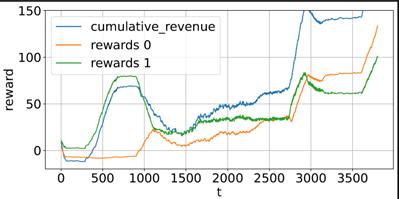}&
\includegraphics[width=0.4\textwidth]{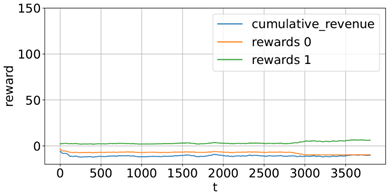}\\
\includegraphics[width=0.4\textwidth]{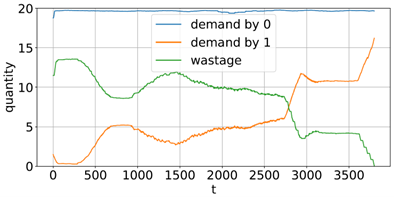}&
\includegraphics[width=0.4\textwidth]{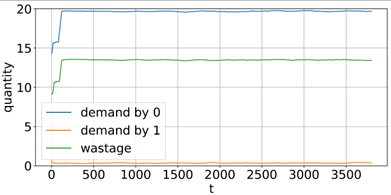}
\end{tabular}
\caption{Additional Results for a Different Price-Demand Setup} \label{fig:additional}
\end{figure*}

Here, we give another empirical evidence in a different price/demand experiment set up (compared to what we used in the main paper) to show that Explicit Data Exchange (EDE) brings significant advantages in cumulative revenue increment and waste minimization compared to No Data Sharing (NDS). Please refer to figure~\ref{fig:additional}. We show graphs for 100 epochs of real data realm followed by initial crude pre-training using simulated data. All external randomness was kept the exact same. Moreover, secure 2 PC framework is computationally equivalent to the EDE realm as we claimed in section 5 of the main paper. \\
The experimental set up for the above diagrams:
Price upper limit of player 0: 10, Price upper limit of player 0: 18, Quantity limit for both of them: 20, Consumer demand dependence on price: $20-0.133p+0.05$(standard normal), Raw material price per unit: 0.5. Normalized Cumulative Revenue = Total Reward – Wastage Cost.\\
So, there is 35.71\% less wastage and 765.91\% better normalized cumulative revenue on average over the entire 100 epochs in Explicit Data Exchange (EDE) compared to No Data Sharing (NDS). Please refer to table~\ref{tab:diff-price-demand-setup} for the details.
\begin{table*}
    \centering
    \begin{tabular}{l|c|c|c|c|c|}
      \toprule 
        &  Avg Reward 0 &  Avg Reward 0 & Total Reward &  Avg Wastage &  Normalized Cumulative Revenue\\      \midrule 
      NDS & -7.38 & 3.33 & -4.05 & 13.27 & -10.68\\
      \hline
      EDE & 31.47 & 43.92 & 75.39 & 8.53 & 71.12 \\
      \bottomrule 
    \end{tabular}
    \caption{Reward, Wastage and Cumulative Revenue for the different price-demand setup.} \label{tab:diff-price-demand-setup}
\end{table*}

\end{document}